\def\year{2021}\relax
\documentclass[letterpaper]{article} 
\usepackage{aaai21}  
\usepackage{times}  
\usepackage{helvet} 
\usepackage{courier}  
\usepackage[hyphens]{url}  
\usepackage{graphicx} 
\def\UrlFont{\rm}  
\usepackage{natbib}  
\usepackage{caption} 
\usepackage[switch]{lineno}  

\usepackage[dvipsnames]{xcolor}
\usepackage{subcaption}
\usepackage{bm}
\usepackage{tabularx}
\newcolumntype{Y}{>{\centering\arraybackslash}X}

\definecolor{darkgreen}{rgb}{0.0, 0.2, 0.13}

\newcommand\todo[1]{\textcolor{red}{(TODO: #1)}}
\newcommand\done[1]{\textcolor{olive}{(DONE: #1)}}

\usepackage{booktabs}

\title{Using Radio Archives for Low-Resource Speech Recognition: \\
Towards an Intelligent Virtual Assistant for Illiterate Users}

\author {
    Moussa Doumbouya,\textsuperscript{\rm 1}
    Lisa Einstein,\textsuperscript{\rm 1,2}
    Chris Piech\textsuperscript{\rm 2}
    \\
}
\affiliations {
    \textsuperscript{\rm 1} GNCode \\
    \textsuperscript{\rm 2} Stanford University \\
    moussa@gncode.org, lisae@stanford.edu, piech@cs.stanford.edu
}

\begin{document}

\maketitle

\begin{abstract}
 For many of the 700 million illiterate people around the world, speech recognition technology could provide a bridge to valuable information and services. Yet, those most in need of this technology are often the most underserved by it. In many countries, illiterate people tend to speak only low-resource languages, for which the datasets necessary for speech technology development are scarce.  In this paper, we investigate the effectiveness of unsupervised speech representation learning on noisy radio broadcasting archives, which are abundant even in low-resource languages. We make three core contributions. First, we release two datasets to the research community. The first, West African Radio Corpus, contains 142 hours of audio in more than 10 languages with a labeled validation subset. The second, West African Virtual Assistant Speech Recognition Corpus, consists of 10K labeled audio clips in four languages. Next, we share West African wav2vec, a speech encoder trained on the noisy radio corpus, and compare it with the baseline Facebook speech encoder trained on six times more data of higher quality. We show that West African wav2vec performs similarly to the baseline on a multilingual speech recognition task, and significantly outperforms the baseline on a West African language identification task. Finally, we share the first-ever speech recognition models for Maninka, Pular and Susu, languages spoken by a combined 10 million people in over seven countries, including six where the majority of the adult population is illiterate. Our contributions offer a path forward for ethical AI research to serve the needs of those most disadvantaged by the digital divide. 
\end{abstract}

\section{Introduction}
Smartphone access has exploded in the Global South, with the potential to increase efficiency; connection; and access to critical health, banking, and education services \cite{mhealth2011new, harris2019mobile, avle2018research}. Yet, the benefits of mobile technology are not accessible to most of the 700 million illiterate people around the world who, beyond simple use cases such as answering a phone call, cannot access functionalities as simple as contact management or text messaging \cite{chipchase2006you}.

Speech recognition technology could help bridge the gap between illiteracy and access to valuable information and services \cite{10.1145/1959022.1959024}, but the development of speech recognition technology requires large annotated datasets. Unfortunately, languages spoken by illiterate people who would most benefit from speech recognition technology tend to fall in the ``low-resource" category, which in contrast with ``high-resource" languages, have few available datasets. Transfer learning, transferring representations learned on high-resource unrelated languages, has not been explored for many low-resource languages \cite{kunze2017transfer}. Even if  transfer learning can help, labeled data is still needed to develop useful models. 

This data deficit persists for multiple reasons. Developing commercial products for languages spoken by smaller populations can be less profitable and thus less prioritized. Furthermore, people with power over technological goods and services tend to speak data-rich languages themselves, potentially leading them to insufficiently consider the needs of users who do not \cite{10.1145/3313831.3376392}.

\begin{figure}
    \centering
    \includegraphics[width=0.95\columnwidth]{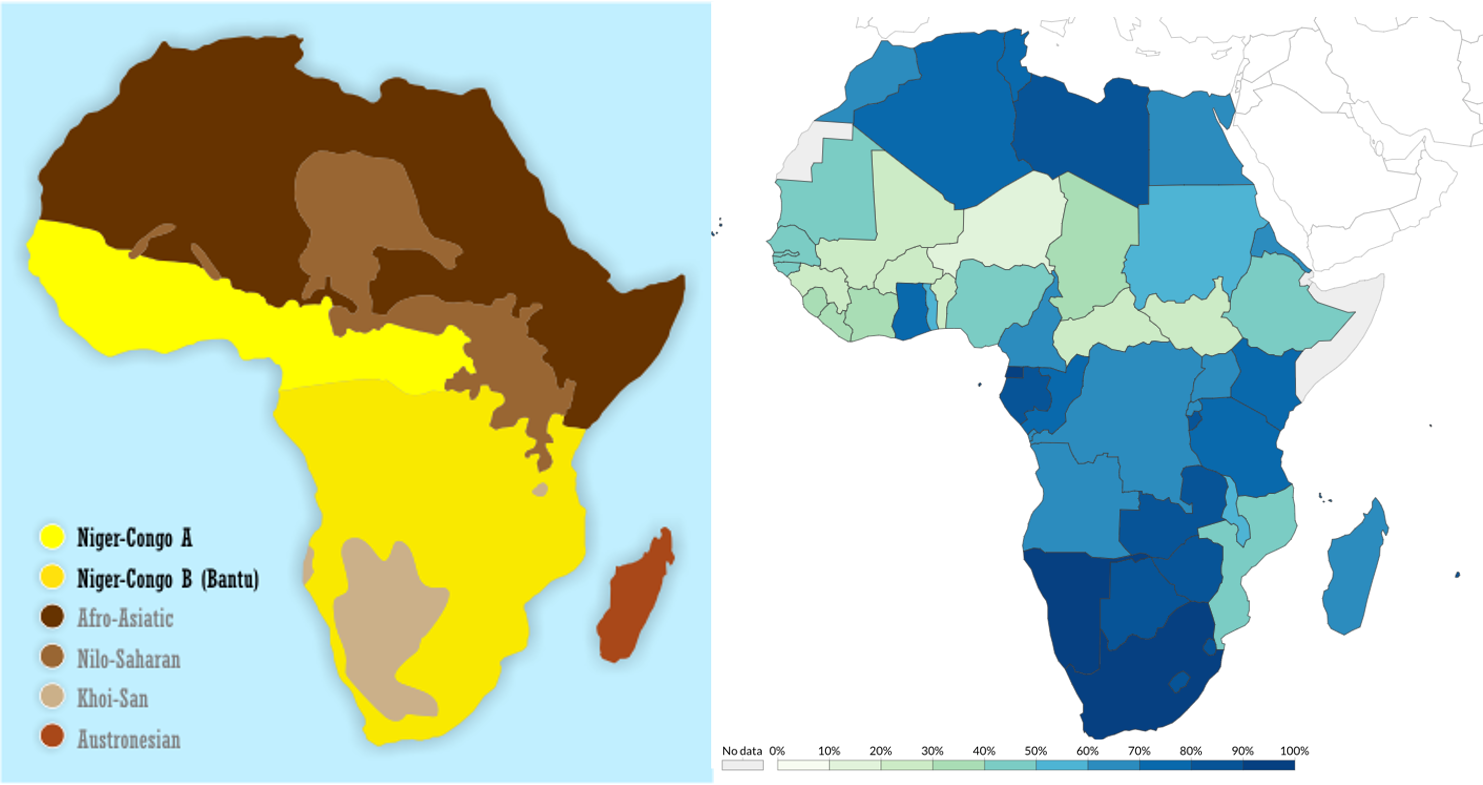}
    \caption{Speaking a language from the Niger Congo family correlates with low literacy rates. Left: Distribution of language families in Africa (``Niger-Congo" by Mark Dingemanse is licensed under CC BY 2.5). Right: Female adult literacy rates in 2015 (Our World in Data/World Bank).}
    \label{fig:adult_illiteracy_rates}
\end{figure}

We take steps toward developing a simple, yet functional intelligent virtual assistant that is capable of contact management skills in Maninka, Susu and Pular, low-resource languages in the Niger Congo family. People who speak Niger Congo languages have among the lowest literacy rates in the world, and illiteracy rates are especially pronounced for women (see Fig. \ref{fig:adult_illiteracy_rates}). Maninka, Pular, and Susu are spoken by a combined 10 million people, primarily in seven African countries, including six where the majority of the adult population is illiterate \cite{owidliteracy}. We address the data scarcity problem by making use of unsupervised speech representation learning and show that representations learned from radio archives, which are abundant in many regions of the world with high illiteracy rates, can be leveraged for speech recognition in low-resource settings.

 \paragraph{In this paper, we make three core contributions that collectively build towards the creation of intelligent virtual assistants for illiterate users:}
 \begin{enumerate}
    \item We present two novel datasets (i) the West African Radio Corpus and (ii) the West African Virtual Assistant Speech Recognition Corpus. These datasets of over 150 hours of speech  increase the availability of resources for speech technology development for West African Languages.
    \item We investigate the effectiveness of unsupervised speech representation learning from noisy radio broadcasting archives. We show that our encoder leads to a multilingual speech recognition accuracy similar to Facebook's baseline state-of-the-art encoder and outperforms the baseline on West African language identification by 13.94\%.
    \item We present the first-ever language identification and small vocabulary speech recognition systems for Maninka, Pular, and Susu. For all languages, we achieve usable performance (88.1\% on automatic speech recognition).
 \end{enumerate}

The results presented are robust enough that our West African speech recognition software, in its current form, is ready to be used effectively in an intelligent virtual assistant capable of contact management skills for illiterate users.
 


The rest of this paper reads as follows. First, we formalize the problem of speech recognition for virtual assistants for illiterate users. Then we provide background information and summarize prior work. We then introduce the novel datasets. Next, we introduce the methodology for our intelligent virtual assistant and present our experimental results. We conclude with a discussion of results and future work. 

\subsection{Contact Management Virtual Assistant}
\label{sec:virtual_assistant_intro}

\begin{figure}
\centering
\includegraphics[width=0.9\columnwidth]{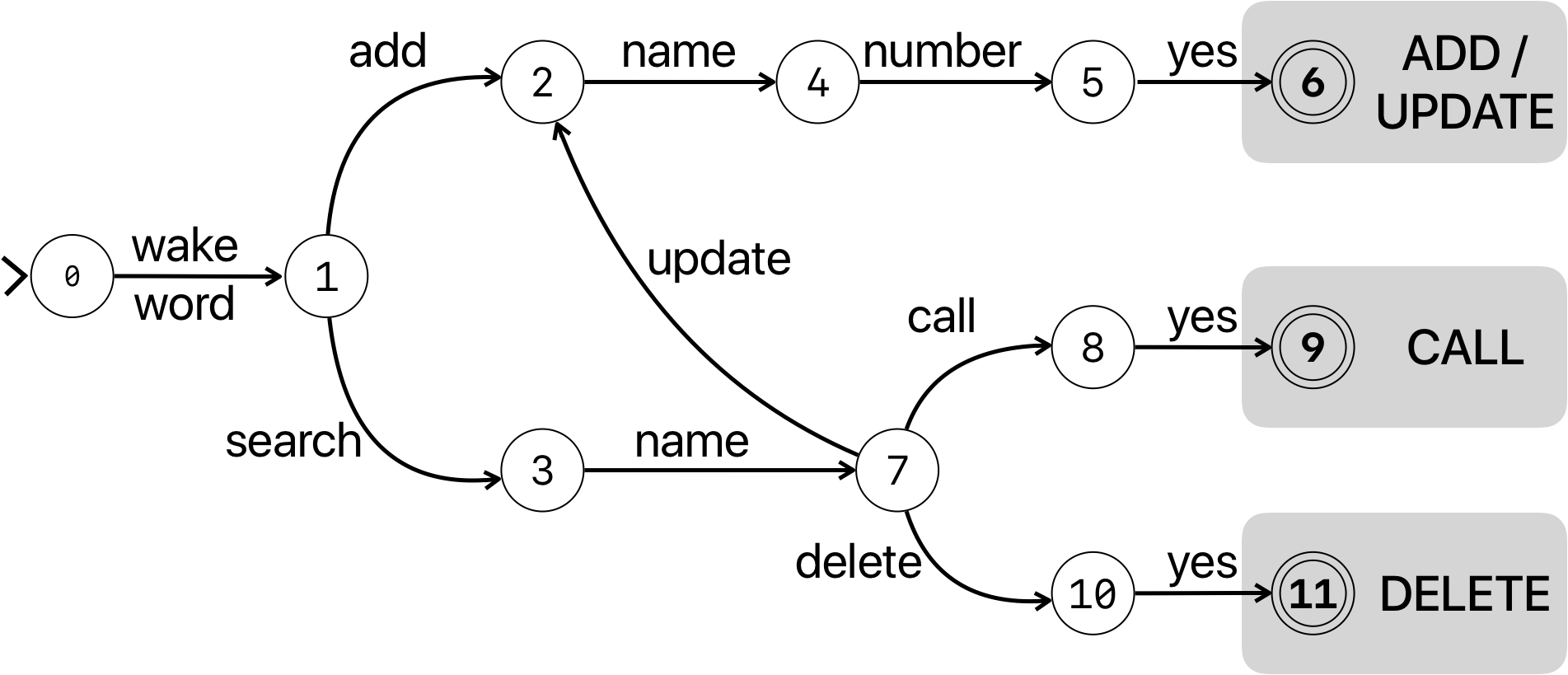} 
\caption{Simplified diagram of the states of the virtual assistant. In non-final states, it recognizes a subset of utterances, e.g., ``add"/``search" in state 1. Upon transition to state 6, 9, or 11, it invokes the contact management application.}
\label{fig:virtual_assistant_states}
\end{figure}

To demonstrate how speech recognition could enable the productive use of technology by illiterate people, we propose a simple yet functional virtual assistant capable of contact management skills in French, Maninka, Susu, and Pular. Fig. \ref{fig:virtual_assistant_states} illustrates the states and transitions of the virtual agent, the performance of which greatly depends on its ability to accurately recognize the user’s utterances.

\paragraph{Automatic Speech Recognition (ASR).} 
As demonstrated in Table \ref{table:dialog_example} and Fig. \ref{fig:virtual_assistant_states}, the recognition of a small utterance vocabulary covering wake words, contact management commands (search, add, update, delete), names, and digits is sufficient to make the assistant functional. 

There are no existing speech recognition systems or data sets for Maninka, Pular, or Susu. Therefore, we first collected and curated the West African Virtual Assistant dataset, which contains the utterances described in Fig. \ref{fig:virtual_assistant_states} in French, Maninka, Pular, and Susu, before creating the speech recognition models.

Because of the small size of our dataset, we used wav2vec, the state of the art unsupervised speech representation learning method from Facebook \cite{schneider2019wav2vec}. We compared the baseline wav2vec model to its counterpart trained on the West African Radio Corpus we collected and conducted West African language identification experiments to validate the learned speech features.

\begin{table}
    \centering
    
    \begin{tabular*}{0.9\columnwidth}{rlc}
        \toprule
        Role & \multicolumn{1}{c}{Utterance} &  State \\ \midrule
        User:& ``Guru!" &  \\
        VA:& ``Yes, what would you like to do?" & 1\\
        \hline
        User:& ``Add a contact" & \\
        VA:& ``What is the name of the contact?" & 2\\
        \hline
        User:& ``Fatoumata" & \\
        VA:& ``What is their phone number?" & 4 \\
        \hline
        User:& ``698332529" & \\
        VA:& ``Are you sure to add Fatoumata" & 5\\
        \hline
        User:& ``Yes" \\
        VA:& ``OK. Done" & 6 \\ 
        \bottomrule
    \end{tabular*}
    
    \caption{Example dialog between a user and the virtual assistant (VA). The last column shows the new state of the VA.}
    \label{table:dialog_example}
\end{table}

\paragraph{Language Identification (Language ID).}
It is not completely clear what happens when speech representations learned from high-resource languages (e.g., English for the baseline wav2vec) are used to solve speech recognition tasks on unrelated low-resource languages such as Maninka, Pular, and Susu. To shed light on the semantics encoded by wav2vec features, we compare the difference in performance on a West African language identification task by the baseline wav2vec with its counterpart trained on the West African Radio Corpus and conduct a qualitative analysis of the acoustic features on which they focus.



\subsection{Prior Work}
\paragraph{User Interfaces for Illiterate People.} Many researchers agree on the importance of facilitating technology access in populations with low literacy rates by using speech recognition and synthesis, local language software, translation, accessibility, and illiterate-friendly software \cite{patra2009ictd, ho2009human, 10.1145/2442882.2442930}. 
 Medhi et al. confirmed illiterate users' inability to use textual interfaces and showed that non-text interfaces significantly outperform their textual counterparts in comparative studies \cite{10.1145/1959022.1959024}. Graphical content and spoken dialog systems have shown promise in allowing illiterate users to perform tasks with their phones or interact with e-government portals \cite{10.1145/1328057.1328125, 10.1145/1959022.1959024, 10.1145/2160601.2160608}. Studies so far have relied on ``Wizard of Oz" voice recognition, where the speech-to-text function is simulated with humans remotely responding to spoken instructions, rather than true ASR as we demonstrate here.  
\paragraph{Existing Speech Datasets for African languages.} Some work has been done to collect speech datasets in low-resource African languages. Documentation exists for three African languages from the Bantu family: Basaa, Myene, and Embosi \cite{Adda2016}. Datasets also exist for other African languages such as Amharic, Swahili, Wolof \cite{Abate2005, Gelas2012, gauthier2016collect}. Several research efforts have focused on South African languages \cite{van-niekerk-etal-2017, Nthite2020,Badenhorst2011}. We were unable to identify any datasets that include the three Niger-Congo languages we focus on here. 

\paragraph{Exploiting ``found" data, including radio broadcasting.}
Cooper et al. explored the use of found data such as ASR data, radio news broadcast, and audiobooks for text-to-speech synthesis for low-resource languages. However, the radio broadcast used is high quality and English language, as opposed to noisy and in low-resource languages \cite{cooper2019text}. Radio Talk, a large-scale corpus of talk radio transcripts, similarly focuses on English language speakers in the United States
\cite{beeferman2019radiotalk}. Some research has focused on speech synthesis from found data in Indian languages \cite{baljekar2018speech,  mendels2015improving}. None of these found data projects include noisy radio data for low-resource languages, a data source that is abundant in many countries with low literacy rates since speech is the method by which citizens must consume information. 

 \paragraph{Unsupervised speech representation learning.} Unsupervised speech representation learning approaches such as  Mockingjay and wav2vec aim to learn speech representation on unlabeled data to increase accuracy on downstream tasks such as phoneme classification, speaker recognition, sentiment analysis, speech recognition, and phoneme recognition while using fewer training data points \cite{Liu_Mockingjay_9054458, schneider2019wav2vec}.

 In this work, we compare the baseline ``wav2vec large" model, trained on  LibriSpeech - a large (960 hours) corpus of English speech read from audiobooks \cite{panayotov2015librispeech} - to its counterpart we trained on a small (142 hours) dataset of noisy radio broadcasting archives in West African languages for the downstream tasks of language identification and speech recognition on West African languages.

\paragraph{Transferring speech representations across languages.}
It has been shown that speech representations learned from a high-resource language may transfer well to tasks on unrelated low-resource languages \cite{riviere2020unsupervised}. In this work, we compare such representations with representations learned on noisy radio broadcasting archives in low-resource languages related to the target languages. We present quantitative results based on performances on downstream tasks, and a qualitative analysis of the encoded acoustic units.


\section{West African Speech Datasets}
In this paper, we present two datasets, the West African Speech Recognition Corpus, useful for creating the speech recognition module of the virtual assistant described in the introduction section, and the West African Radio Corpus intended for unsupervised speech representation learning for downstream tasks targeting West African languages.

\subsection{West African Virtual Assistant Speech Recognition Corpus}
The West African Speech Recognition Corpus contains 10,083 recorded utterances from 49 speakers (16 female and 33 male) ranging from 5 to 76 years old on various devices. Most speakers are multi-lingual and were recorded in all languages they spoke. First names were recorded once per speaker, as they are language independent.  

Following the virtual assistant model illustrated in Fig. \ref{fig:virtual_assistant_states}, the ASR corpus consists of the following utterances in French, Maninka, Susu, and Pular: a wake word, 7 voice commands (``add a person", ``search a person", ``call that", ``update that", ``delete that", ``yes", ``no"), 10 digits, ``mom" and ``dad". The corpus also contains 25 popular Guinean first names useful for associating names with contacts in a small vocabulary speech recognition context. In total, the corpus contains 105 distinct utterance classes.

82\% of the recording sessions were performed simultaneously on 3 devices (one laptop, and two smartphones). This enables the creation of acoustic models that are invariant to device-specific characteristics and the study of sensitivity with respect to those characteristics.

Each audio clip is annotated with the following fields: Recording session, speaker, device, language, utterance category (e.g., add a contact), utterance (e.g., add a contact in Susu language), and the speaker's age, gender, and native language. Speakers have been anonymized to protect privacy. Table \ref{table:asr_corpus_stats} (top) summarizes the content of the West African Virtual Assistant Speech Recognition Corpus.

\begin{table}[t]
    \centering
    
    \resizebox{\columnwidth}{!}{
    \begin{tabular}{lrrrr}
    \multicolumn{5}{c}{\textbf{Dataset 1}}\\
    \multicolumn{5}{c}{\textbf{West African Virtual Assistant Speech Recognition Corpus}}\\
     \toprule
        Utterance Category & French & Maninka & Pular & Susu \\ \midrule
        Wake word & 66 & 95 & 67 & 109 \\ 
        Add  & 66 & 95 & 67 & 111 \\ 
        Search & 66 & 95 & 67 & 111 \\ 
        Update & 66 & 95 & 67 & 111 \\ 
        Delete & 66 & 95 & 67 & 111 \\ 
        Call & 66 & 95 & 64 & 111 \\ 
        Yes & 66 & 95 & 67 & 111 \\ 
        No & 66 & 95 & 67 & 111 \\ 
        Digits (10) & 660 & 946 & 670 & 1,110 \\ 
        Mom & 36 & 53 & 43 & 51 \\ 
        Dad & 36 & 53 & 43 & 51 \\ \midrule
        \textbf{Total/Language} & 1,260 & 1,812 & 1,289 & 2,098 \\ \midrule
        Names (25) & \multicolumn{4}{c}{3,624} \\ \midrule
        \textbf{Total} & \multicolumn{4}{c}{10,083} \\ \bottomrule
        \\
        \\
        \\
        
    \multicolumn{5}{c}{\textbf{Dataset 2}}\\
    \multicolumn{5}{c}{\textbf{West African Radio Corpus}}\\
    \toprule
    & \multicolumn{2}{r}{Clips} & \multicolumn{2}{r}{Duration} \\
    \midrule
    Unlabeled Set & \multicolumn{2}{r}{17,091} & \multicolumn{2}{r}{142.4 hours} \\
    Validation Set & \multicolumn{2}{r}{300} & \multicolumn{2}{r}{2.5 hours} \\ \hline
    \textbf{Total} & \multicolumn{2}{r}{17,391} & \multicolumn{2}{r}{144.9 hours} \\ \bottomrule
    \\
    \\

        
    \end{tabular}
    } 
    
    \caption{Description of collected datasets. Dataset 1: Record counts by utterance category in the West African Virtual Assistant Speech recognition corpus. We aggregated record counts for digits (10 per language) and names (25 common Guinean names). Dataset 2: West African Radio Corpus includes noisy audio in over 10 local languages and French collected from six Guinean radio stations.}
    \label{table:asr_corpus_stats}
\end{table}

\subsection{West African Radio Corpus}
The West African Radio Corpus consists of 17,091 audio clips of length 30 seconds sampled from archives collected from six Guinean radio stations. The broadcasts consist of news and various radio shows in languages including French, Guerze, Koniaka, Kissi, Kono, Maninka, Mano, Pular, Susu, and Toma. Some radio shows include phone calls, background and foreground music, and various noise types.  Although an effort was made to filter out archive files that mostly contained music, the filtering was not exhaustive. Therefore, this dataset should be considered uncurated. Segments of length 30 seconds were randomly sampled from each raw archive file. The number of sampled segments was proportional to the length of the original archive, and amounts to approximately 20\% of its length.

The corpus also contains a validation set of 300 audio clips independently sampled from the same raw radio archives, but not included in the main corpus. The validation clips are annotated with a variety of tags including languages spoken, the presence of a single or multiple speakers, the presence of verbal nods, telephone speech, foreground noise, and background noise among other characteristics.

\section{Method}
\subsection{West African wav2vec (WAwav2vec)}
To maintain comparability with wav2vec, WAwav2vec was obtained by training wav2vec as implemented in the fairseq framework \cite{ott2019fairseq} on the West African Radio Corpus. We used the ``wav2vec large” model variant described in \cite{schneider2019wav2vec} and applied the same hyperparameters, but we trained on 2 Nvidia GTX 1080 Ti GPUs instead of 16 GPUs as did \cite{schneider2019wav2vec}. We trained for 200k iterations (170 epochs) and selected the best checkpoint based on the cross-validation loss. The audio clips from the West African Radio Corpus were converted to mono channel waveforms with a sampling rate of 16 kHz and normalized sound levels.
The baseline wav2vec and the WAwav2vec were used as feature extractors in all our experiments. We experimented with both their context (C) and latent (Z) features. We used quantitative and qualitative observations on the downstream tasks and analysis to make conclusions about the effectiveness of unsupervised speech representation learning and transfer learning in two settings: The first, where representations are learned from high-quality large-scale datasets in a high-resource language not directly related to the target languages, and the second, where representations are learned from noisy radio broadcasting archives in languages related to target languages.


\subsection{Neural Net for Virtual Assistant}
\begin{figure}
\centering
\includegraphics[width=1.0\columnwidth]{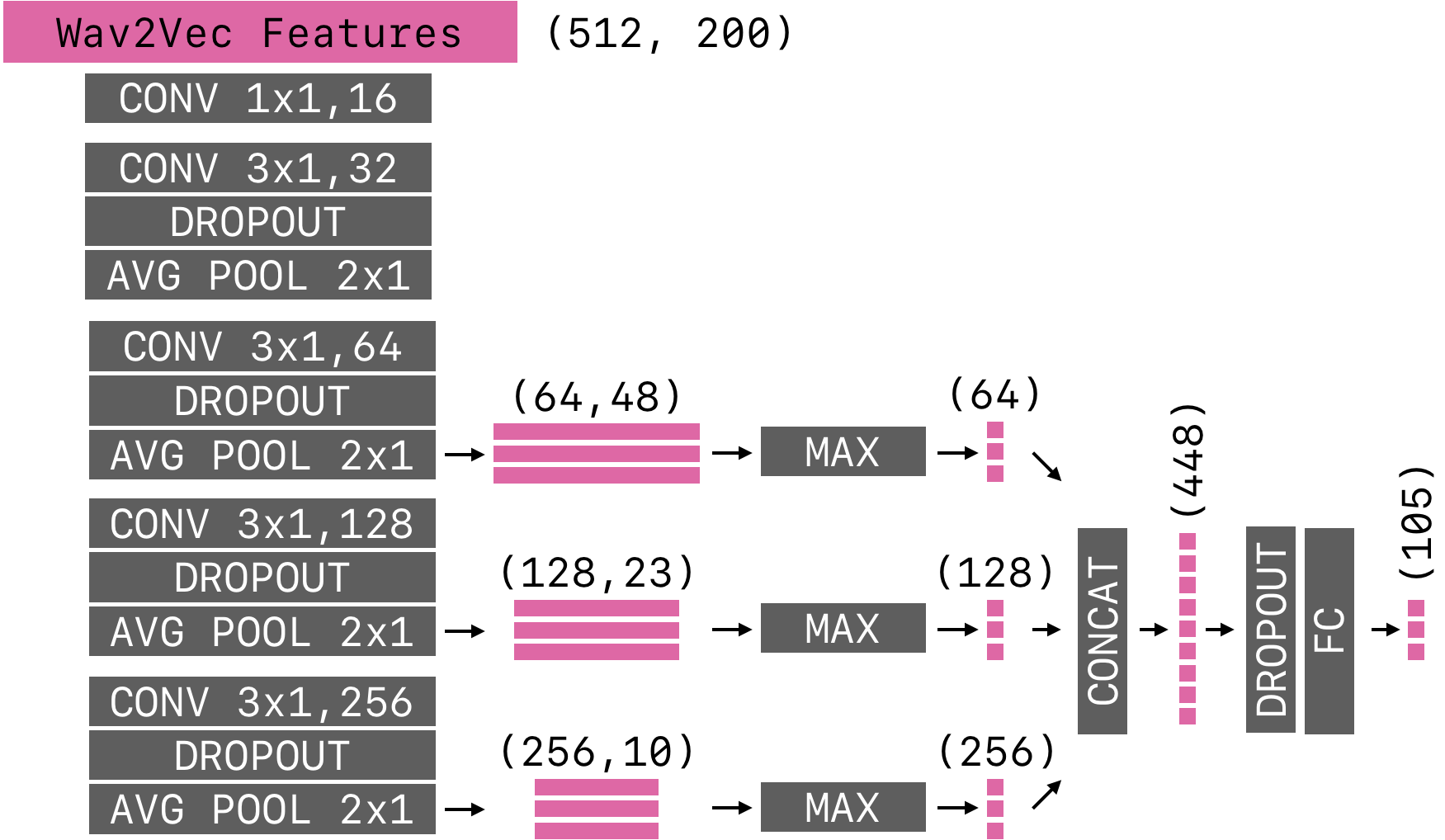} 
\caption{Architecture of the speech recognition CNN. The language identification CNN has a similar architecture. }
\label{fig:asrcnn3_architecture}
\end{figure}

We used the convolutional neural network architecture illustrated in Fig. \ref{fig:asrcnn3_architecture} for both the language identification and the speech recognition experiments. Of its variants we explored, the following performed the best. The network comprises a 1x1 convolution followed by 4 feature extraction blocks. Each feature extraction block contains a 3x1 convolution, the ELU activation function \cite{clevert2015fast}, a Dropout layer \cite{srivastava2014dropout} and an average pooling layer with kernel size 2 and stride 2. The output of the last 3 feature extraction blocks are max pooled across the temporal dimension and then concatenated to make a fixed-length feature vector that is fed to the fully connected layer. This design allows extracting acoustic features at multiple scales and makes the neural network applicable to any sequence length. In order to mitigate overfitting issues, we apply Dropout in each of the convolutions feature extractors and before the fully connected layer.

\begin{table}
    \centering
        \begin{tabularx}{0.8\columnwidth}{ccc}
            \toprule 
            Model & wav2vec & mel spectrogram \\ \midrule
            Language ID & 1,651 & 499 \\ 
            ASR & 186,393 & 180,249 \\ \bottomrule
        \end{tabularx}
    \caption{Parameter counts of the CNNs for language identification and speech recognition using wav2vec features and mel spectrograms (respectively 512 and 128 dimensional).}
    \label{tab:parameter_counts}
\end{table}

The language identification model uses 3, 1, 3, 3 and 3 convolution channels, resulting in a 9 dimensional feature vector used for a 3 class classification. The ASR model uses 16, 32, 64, 128 and 256 convolutional chanels, resulting in a 448 dimensional representation used for a 105 class classification.  In both experiments, we used the Adam optimiser \cite{kingma2014adam} with learning rate $10^{-3}$ to minimize a cross entropy loss function. We also compared learned wav2vec features with spectrograms based on 128 mel filter banks.
Table \ref{tab:parameter_counts} summarizes the number of parameters of each of our models. We implemented and trained our speech recognition and language identification networks using PyTorch  \cite{paszke2019pytorch}.

\section{Results}
In addition to establishing the baseline accuracies for speech recognition on the West African Virtual Assistant Speech Recognition Corpus and language identification on the validation set of the West African Radio Corpus, our experiments aimed at answering the following questions:
\begin{itemize}
    \item Is it possible to learn useful speech representations from the West African Radio Corpus?
    \item How do such representations compare with the features of the baseline wav2vec encoder, trained on a high-quality large-scale English dataset, for downstream tasks on West African languages? 
    \item How does the West African wav2vec qualitatively compare with the baseline wav2vec encoder?
\end{itemize}

\subsection{Language Identification}
We used the annotated validation set of the West African Radio Corpus, which is disjoint from its unlabeled portion on which WAwav2vec is trained, to train the language identification neural network for the task of classifying audio clips in Maninka, Pular, and Susu.

We selected clips where the spoken languages include exactly one of Maninka, Susu, or Pular. For balance, we selected 28 clips per language for a total of 84 clips. Because of the small data size, we performed 10-fold cross-validation with randomly sampled training (60\%) and validation (40\%) portions. The mean test accuracies and their standard errors are reported in Table \ref{table:langid_results_summary}, showing that the West African wav2vec features outperform the baseline wav2vec, which outperforms mel spectrograms. 

\begin{figure}
    \centering

    \includegraphics[width=0.9\columnwidth]{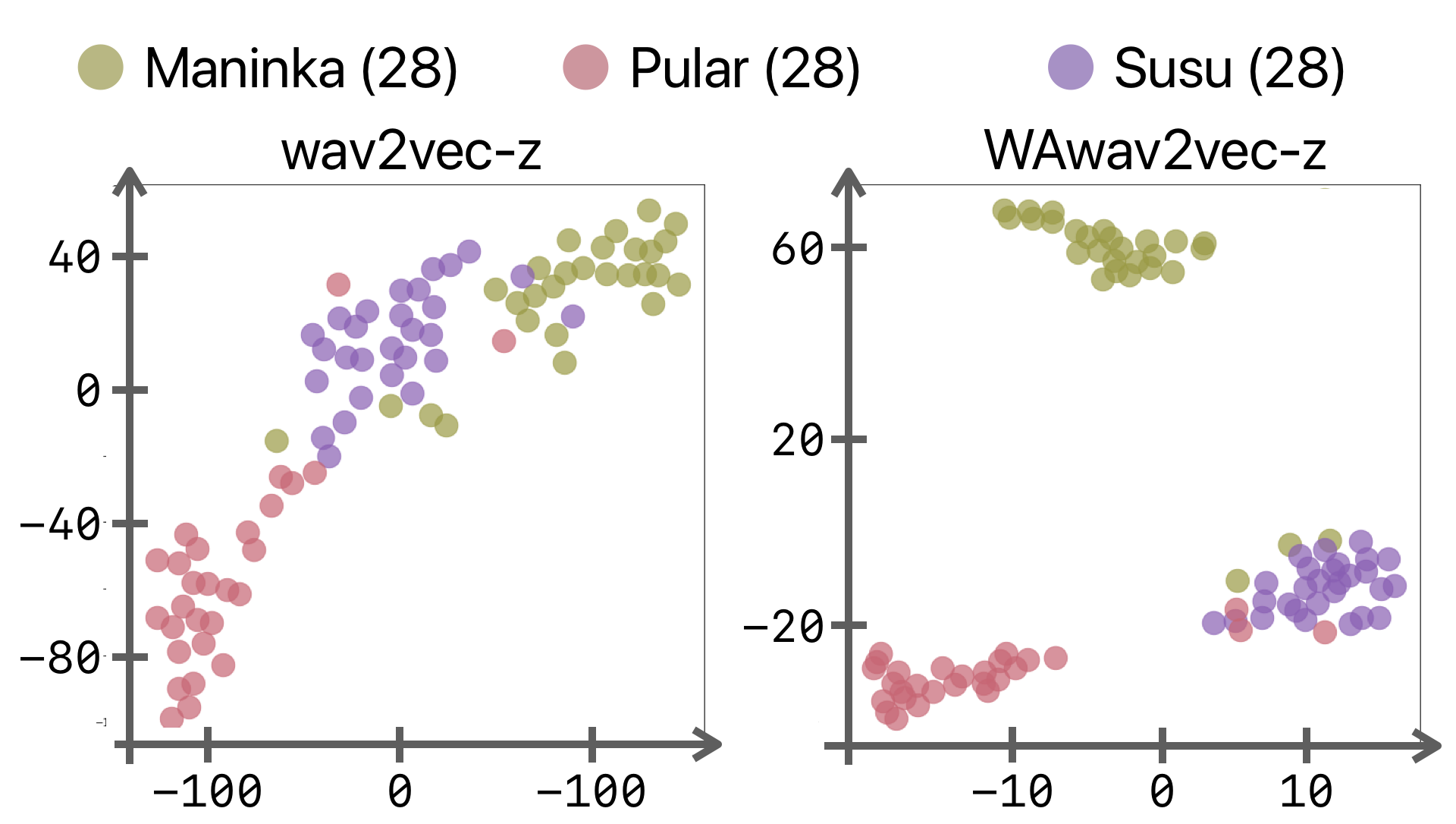}
    \caption{t-SNE projection of the language ID CNN's intermediate features for 84 audio clips encoded using the baseline wav2vec (left), and the West African wav2vec (right).}
    \label{fig:langid_ensemble_clip_representation_tsne}
\end{figure}

Fig. \ref{fig:langid_ensemble_clip_representation_tsne} shows the 84 audio clips used in the language identification experiments. The aggregated concatenated 9-D convolutional features of the best model for each of the 10 cross-validation training sessions were concatenated to make 90-D feature vectors. As bolstered by the qualitative results in the \textit{Acoustic Unit Segmentation} section, the t-SNE \cite{maaten2008visualizing} projection of those feature vectors suggests that the WAwav2vec encoder is more sensitive than the baseline wav2vec to the specificities of the Maninka, Susu, and Pular languages.

\begin{table}
    \centering
    \resizebox{\columnwidth}{!}{
    \begin{tabular}{ccccc}
    \toprule
         Features & \multicolumn{4}{c}{Test Accuracy} \\ \midrule
          & Overall & MA & PU & SU \\ \hline
        mel spectrogram & $60.00 \pm 2.80$  & $60.79$ & $72.93$ & $40.55$ \\ \hline
        
        wav2vec-z & $65.15\pm2.20$  & $58.05$ & $78.97$ & $60.08$ \\
        WAwav2vec-z & $\bm{79.09 \pm 1.32}$  & $\bm{88.43}$ & $83.76$ & $63.44$ \\ \hline
        
        wav2vec-c & $60.61 \pm 0.61$  & $70.31$ & $72.43$ & $34.61$ \\
        WAwav2vec-c & $78.48 \pm 1.51$  & $80.09$ & $\bm{88.20}$ & $\bm{67.02}$ \\ \bottomrule
    \end{tabular}
   }
    \caption{Language ID accuracies using mel spectrograms, and the latent (z) and context (c) features of the baseline wav2vec, and the West African wav2vec (WAwav2vec).} 
    \label{table:langid_results_summary}
\end{table}


\subsection{Multilingual Speech Recognition}
Next, we compared WAwav2vec to the baseline wav2vec encoder for the downstream task of speech recognition on the West African Virtual Assistant Speech Recognition Corpus containing 105 distinct utterance classes across 4 languages.
Table \ref{table:asr_results} summarizes the speech recognition accuracies, from which we conclude that the features of the West African wav2vec are on par with the baseline wav2vec for the task of multilingual speech recognition.

\begin{table*}[ht]
    \centering
    \resizebox{\textwidth}{!}{
        \begin{tabular}{cccccccc}
        \toprule
             Features & \multicolumn{7}{c}{Test Accuracy} \\ \midrule
             & Overall & Names &French & Maninka & Pular & Susu & Native Language\\ \midrule
            mel spectrogram & 
                $74.05 \pm 0.74$ & 
                $75.88 \pm 0.74$ & 
                $67.79 \pm 1.98$ & 
                $73.37 \pm 0.56$ &
                $71.60 \pm 1.78$ &
                $78.59 \pm 1.51$ &
                $71.75 \pm 0.97$ \\ \hline
                
            wav2vec-z & 
                $88.36 \pm 0.45$ & 
                $89.41 \pm 0.75$ & 
                $85.44 \pm 1.32$ & 
                $86.80 \pm 0.96$ & 
                $\bm{91.59 \pm 0.79}$ & 
                $88.27 \pm 0.95$ &
                $87.73 \pm 0.39$ \\
                
            WAwav2vec-z & 
                $87.64 \pm 0.63$ &
                $89.59 \pm 1.21$ &
                $83.27 \pm 0.84$ &
                $85.92 \pm 0.81$ &
                $87.72 \pm 1.24$ &
                $89.74 \pm 0.53$ &
                $86.49 \pm 0.40$ \\ \hline
                
            wav2vec-c & 
                $\bm{88.79 \pm 0.46}$ &
                $\bm{89.78 \pm 0.48}$ &
                $\bm{86.92 \pm 1.38}$ &
                $87.51 \pm 0.96$ &
                $88.75 \pm 0.72$ &
                $\bm{89.88 \pm 1.11}$ &
                $87.54 \pm 0.40$ \\ 
                
            WAwav2vec-c &
                $88.01 \pm 0.43$ &
                $87.74 \pm 1.10$ &
                $84.50 \pm 1.03$ &
                $\bm{88.53 \pm 0.00}$ &
                $89.19 \pm 1.27$ &
                $89.59 \pm 0.40$ &
                $\bm{87.99 \pm 0.53}$ \\ 
            \bottomrule
                
        \end{tabular}
    }
    \caption{Multilingual ASR Accuracies on the West African Virtual Assistant Speech  Recognition Corpus: overall, for Guinean names, for utterances in specific languages, and for utterances spoken in the native language of the speaker. We compare models using mel spectrograms, the latent (z) and context (c) features of the baseline wav2vec, and those of WAwav2vec.}
    \label{table:asr_results}
\end{table*}

\subsection{Acoustic Unit Segmentation}
The previous experimental results showed that while features of the baseline wav2vec were overall marginally better than those of WAwav2vec for multilingual speech recognition, the features of WAwav2vec outperformed the baseline on the task of West African language identification. In this section, we attempt to qualitatively analyse the nature of the salient acoustic units encoded by both speech encoders.

\begin{figure}[ht!]
    \centering
    
    \begin{subfigure}[b]{\columnwidth}
        \centering
        \includegraphics[width=0.9\columnwidth]{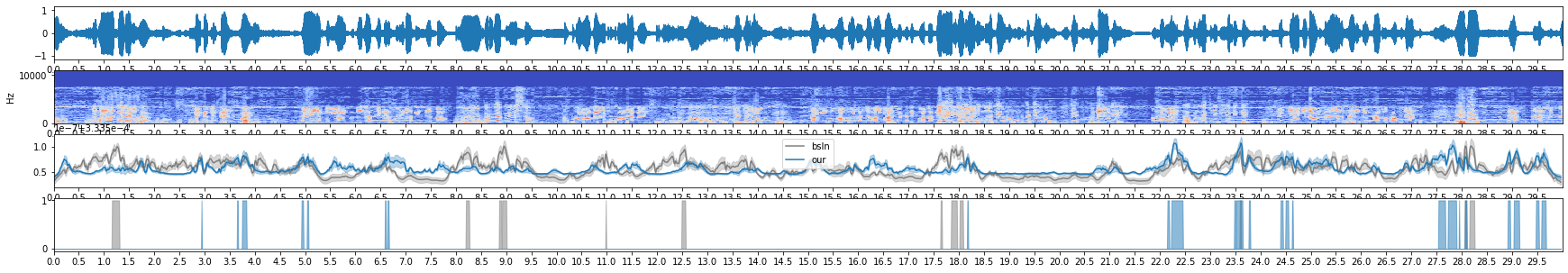}
        \caption{Raw audio, attention signal, and segmented acoustic units.}
        \label{fig:units:attention_signal}
    \end{subfigure}

    \begin{subfigure}[b]{\columnwidth}
        \centering
        
        \includegraphics[width=0.9\columnwidth]{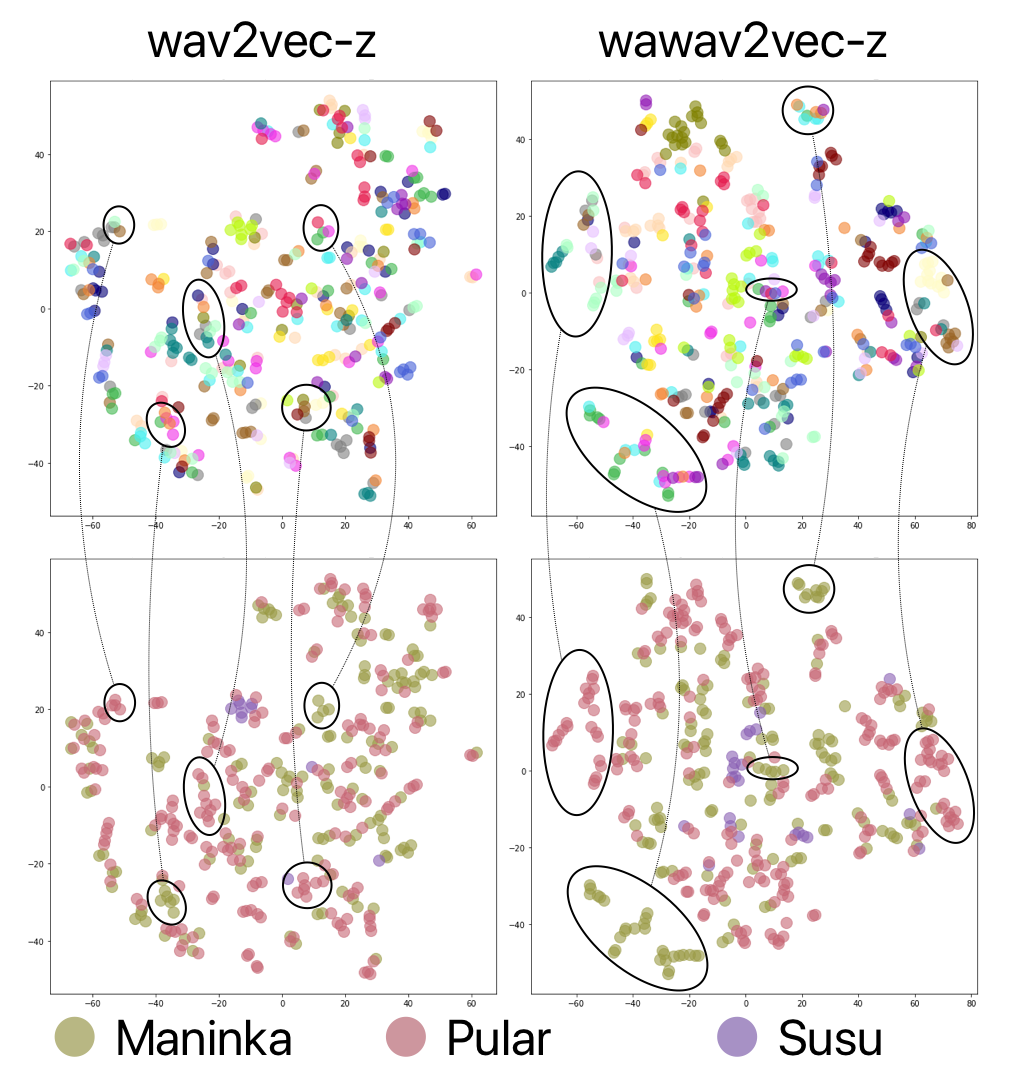}
        \caption{t-SNE Projection of segmented acoustic units. Left: baseline wav2vec. Right: West African wav2vec}
        \label{fig:units:tsne_projection}
    \end{subfigure}

    \begin{subfigure}[b]{\columnwidth}
        \centering
        \includegraphics[width=0.9\columnwidth]{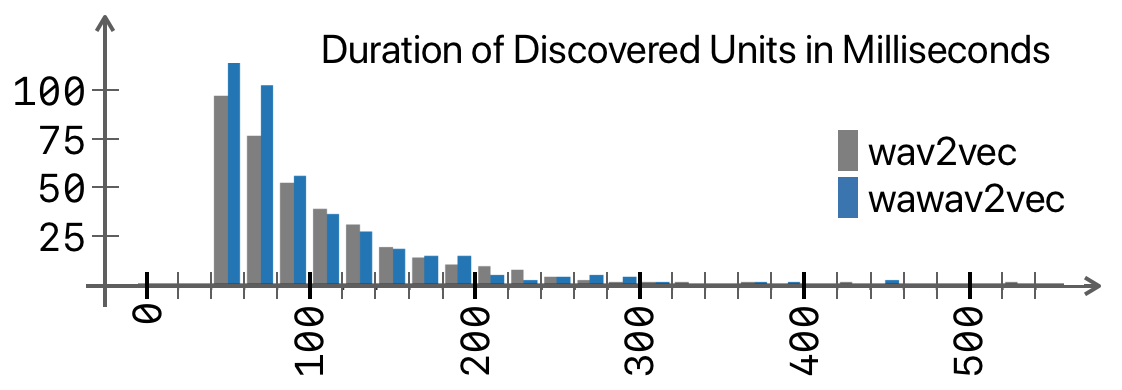}
        \caption{Histograms of the duration of the acoustic units  in milliseconds.}
        \label{fig:units:duration_histogram}
    \end{subfigure}

    \caption{Visualization of segmented acoustic units. (\protect\subref{fig:units:attention_signal}) First two rows: raw wave form and spectrogram of a classified audio clip. Last two rows: attention signals and segmented units computed using wav2vec and WAwav2vec. (\protect\subref{fig:units:tsne_projection}) t-SNE projection of segmented acoustic units colored by audio clip (top row) and language (bottom row) using wav2vec features (left) and WAwav2vec features (right) with highlighted clusters of acoustic units of the same language extracted from different audio clips. (\protect\subref{fig:units:duration_histogram}) Histogram of the duration of the segmented acoustic units in milliseconds.
    }
    
    \label{fig:units}
\end{figure}

We identified important acoustic segments that influence the language classification decision by computing the gradients of the input features with respect to the output of the language identification neural network, similarly to \cite{simonyan2013deep}, but with speech instead of images. We computed an attention signal by first normalizing (using softmax) the magnitude of the gradients, then summing them across the 512 input features, and finally normalizing again over the input sequence. Fig. \ref{fig:units:attention_signal} shows the attention signal aligned with the audio waveform.

Important acoustic units were segmented by considering contiguous time steps where the attention signal was 2 standard deviations above its mean. Fig. \ref{fig:units:duration_histogram} shows the histogram of the duration $d = (c - 1) \times p + f$ of the segmented acoustic units computed as a function of the number of wav2vec time steps ($c$), the period of the wav2vec latent features $p=10 ms$, and their receptive field with respect to the raw input waveform $f=30 ms$.

To get a sense of the nature of the segmented acoustic units, we computed their representation by averaging the 512-D wav2vec features over their span, and then projected those features to 2-D using t-SNE.  Fig. \ref{fig:units:tsne_projection} shows that the acoustic units in WAwav2vec features tend to separate by language more than the ones in the baseline wav2vec.

The duration of the segmented acoustic units and their language-specificity hint that they might be phoneme-like. In future work, we would like to investigate this approach to phoneme discovery, which to the best of our knowledge has not been explored before.

Fig. \ref{fig:units} reveals neighborhoods of 40-200 milliseconds long language-specific acoustic units particularly prominent in the features of WAwav2vec, suggesting the relevance of its encoding to Maninka, Pular, and Susu.


\section{Discussion}

\paragraph{We developed the first-ever speech recognition models for Maninka, Pular and Susu.}
To the best of our knowledge, the multilingual speech recognition models we trained are the first-ever to recognize speech in Maninka, Pular, and Susu. We also showed how this model can power a voice interface for contact management.

\paragraph{We enabled a multilingual intelligent virtual assistant for three languages spoken by 10 million people in regions with low literacy rates.}
The state diagram shown in Fig. \ref{fig:virtual_assistant_states} demonstrates that the virtual assistant is simple yet functional and usable for contact management, provided an ASR model capable of recognizing the utterances described in Table \ref{table:asr_corpus_stats}. We built a speech recognition model capable of classifying those utterances with more than $88\%$ accuracy. We expect good generalization performance given the diversity of devices used for data collection, and the low variance of accuracy across the validation folds. The virtual assistant has a distinct wake word for each language. Therefore, after activation, it only needs to recognize utterances in the language corresponding to the used wake word. Additionally, as Fig. \ref{fig:virtual_assistant_states} shows, at each state there is only a subset of the utterance vocabulary that the assistant needs to recognize. Consequently, in practice the virtual assistant's speech recognition accuracy will be above the accuracy reported in our experiments.

\paragraph{Noisy radio archives are useful for unsupervised speech representation learning in low-resource languages.}
WAwav2vec features significantly improved over mel spectrograms in both ASR accuracy ($88.01\%$ vs $74.05\%$) and language ID accuracy ($79.09\%$ vs $60.00\%$).

\paragraph{WAwav2vec is on par with wav2vec on multilingual speech recognition.}
Speech features learned from the West African Radio corpus lead to $88.01\%$ speech recognition accuracy, which is on par with the accuracy obtained with the baseline wav2vec, $88.79\%$. This result may be surprising given that the radio corpus is of lower quality (noise, multi-speakers, telephone, background and foreground music, etc.), and smaller size (142 vs 960 hours) compared to LibriSpeech, the training dataset of the baseline wav2vec. However, this result may be justified because the languages spoken in the West African Radio Corpus are more closely related to the target languages compared to English.

\paragraph{WAwav2vec outperforms wav2vec on West African Language Identification.} On the task of language identification, WAwav2vec features outperformed the baseline by a large margin, achieving $79.09\%$ accuracy compared to the baseline accuracy of $65.15\%$. Our qualitative analysis indicated that the language classifier's decision was influenced by acoustic units of duration 40 to 200 milliseconds. Data visualization suggested that the acoustic units segmented from WAwav2vec features were more language-specific than the ones segmented from the baseline wav2vec features.

\paragraph{English speech features can be useful for speech recognition in West African languages.} Using the baseline wav2vec resulted in $88.79\%$ speech recognition accuracy, compared to $74.05\%$ with mel spectrograms.

\paragraph{There are non-obvious trade-offs for unsupervised speech representation learning.} WAwav2vec performs as well as the baseline wav2vec on the task of multilingual speech recognition, and outperforms the baseline wav2vec on West African language identification. This indicates the need for a more rigorous investigation of the trade-offs between relevance, size and quality of datasets used for unsupervised speech representation learning.

\paragraph{We publicly released useful resources for West African speech technology development.} 
To advance speech technology for West African languages we released the West African Radio Corpus \footnote{https://openslr.org/105}, the West African Virtual Assistant Speech Recognition Corpus  \footnote{https://openslr.org/106}, and a prototype of our multi-lingual intelligent virtual assistant along with our trained models and code to reproduce our experiments.\footnote{https://github.com/mdoumbouya/nicolingua}

\section{Limitations and Future Work}
The virtual assistant only recognizes a limited vocabulary for contact management. Future work could expand its vocabulary to application domains such as micro-finance, agriculture, or education. We also hope to expand its capabilities to more languages from the Niger-Congo family and beyond, so that literacy or ability to speak a foreign language are not prerequisites for accessing the benefits of technology. The abundance of radio data should make it straightforward to extend the encoder to other languages. Also, training on more languages and language families (e.g., Mande and Bantu languages) might lead to higher accuracy. In our results, the West African wav2vec found acoustic units highly correlated with languages in our dataset. This hints at the potential use of speech encoders to learn language-specific linguistic features. We have only scratched the surface of using unsupervised speech representation learning to better articulate what makes each language unique. 





\section{Conclusion}
We introduced a simple, yet functional virtual assistant capable of contact management for illiterate speakers of Maninka, Pular, and Susu, collected the dataset required to develop its speech recognition module, and established baseline speech recognition accuracy.

To address the low-resource challenge, we explored unsupervised speech representation learning in two contexts. First, where representations are learned from a high-resource language unrelated to the target low-resource languages. Second, where representations are learned from low-quality radio archives in languages related to the target low-resource languages. We gathered quantitative comparative results, developed an effective qualitative analysis method of the learned representations, and showed the benefit of learning speech representations from radio archives, which are abundant even in low-resource languages.

We created the first-ever speech recognition models for three West African languages. We also publicly released all our developed software, trained models, and collected datasets to promote further speech technology development for currently marginalized communities.

\section{Acknowledgements}
We thank Voix de Fria, Radio Rurale de Fria, Radio Rurale Regionale de N'Zerekore, Radio Baobab N'Zerekore, City FM Conakry, and GuiGui FM Conakry for donating radio archives and the 49 speakers who contributed their voices and time to the speech recognition corpus. We thank Taibou Camara for annotating and collecting data, with support from Moussa Thomas Doumbouya, Sere Moussa Doumbouya and Djene Camara. Koumba Salifou Soumah, Mamadou Alimou Balde and Mamadou Sow provided valuable input on the vocabulary of the virtual assistant and Youssouf Sagno, Demba Doumbouya, Salou Nabe, Djoume Sangare and Ibrahima Doumbouya supported various aspects of the project including mediating with radio stations. We thank FASeF for providing office space and relatively stable grid power essential to running our deep learning experiments. Koulako Camara provided critical resources and guidance that made this project possible.
\\
\section{Ethics Statement}
\textbf{Social Justice \& Race}  It is well known that digital technologies can have different consequences for people of different races \cite{10.1145/2851581.2892578}. Technological systems can fail to provide the same quality of services for diverse users, treating some groups as if they do not exist \cite{10.1145/3313831.3376445}. Speakers of West African low-resource languages are likely to be ignored given that they are grossly underrepresented in research labs, companies and universities that have historically developed speech-recognition technologies. Our work serves to lessen that digital divide, with intellectual contributions, our personal backgrounds, and the access to technology we seek to provide to historically marginalized communities. \\
\textbf{Researchers} This research was conducted by researchers raised in Guinea, Kenya, Malaysia, and the United States. Team members have extensive experience living and working in Guinea, where a majority of this research was done, in collaboration with family members, close friends, and the local community. \\
\textbf{Participants} All humans who participated in data creation in various languages were adults who volunteered and are aware of and interested in the impact of this work. \\
\textbf{Data} The data contains the ages, genders and native languages of participants, but names have been erased for anonymity.\\
\textbf{Copyright} Radio data is being made public with permission from the copyright holders. \\
\textbf{Finance} The authors are not employed by any company working to monetize research in the region. \\

\bibliography{main.bib}

\end{document}



\maketitle

%
%

\onecolumn


\section{Appendix A: Manifest of Supplementary Materials Packages}
\pagebreak

\begin{table}
    \centering
    \begin{tabular}{ll}
        \multicolumn{2}{c}{\textbf{Manifest of supplementary materials package: CodeAndDataAppendix}} \\
        \toprule
        Path & Description  \\
        \midrule
        \quad \verb|lang_id_and_unit_segmentation/| & Artifacts for language ID and acoustic unit segmentation experiments \\
        \quad \quad \verb|audio_segments/|  & 10 audio segments features in jupyter notebook \\
        \quad \quad \verb|wav2vec_features/|  & wav2vec features of the 10 audio segments \\
        \quad \quad \quad \verb|wav2vec_features-z/|  & Baseline Wav2vec Latent  \\
        \quad \quad \quad \verb|retrained-wav2vec_features-z/|  & West African Wav2vec Latent \\
        \quad \quad \verb|audacity_markers/|  & Audacity marker files with segmented acoustic units \\
        \quad \quad \verb|results_209/|  & Raw results and log files for 10 folds Language ID training trials \\
        \quad \quad \verb|lang_id_and_unit_seg_results.html|  & Jupyter notebook snapshot for Lang ID and unit segmentation results \\
        \quad \quad \verb|209_gn_lang_id_cnn_training.html|  & Jupyter notebook snapshot for Lang ID CNN Training \\
        \midrule
        \quad \verb|va_asr/|  &  Code For training the Speech Recognition Model \\
        \quad \quad \verb|data.py|  &  Data Loading Utilities\\
        \quad \quad \verb|models.py|  &  Definition of neural networks \\
        \quad \quad \verb|results.py|  &  Utilities for writing results \\
        \quad \quad \verb|train_va_asr.py|  &  Training code\\
        \quad \quad \verb|utils.py|  &  Misc utilities\\
        \quad \quad \verb|requirements.txt|  &  Snapshot of python environment\\
        \quad \quad \verb|Makefile|  &  Entry points for ASR training with Mel Fbanks and Wav2vec features\\
        \quad \quad \verb|results_306/|  &  Raw results and log files for 5 folds Speech Recognition training trials\\
        \quad \quad \verb|asr_results.html|  & Jupyter notebook snapshot for ASR results \\
        \bottomrule
    \end{tabular}
    
    \caption{Manifest of CodeAndDataAppendix}
    \label{table:manifest_code_and_data_appendix}
\end{table}

\begin{table}
    \centering
    \begin{tabular}{ll}
        \multicolumn{2}{c}{\textbf{Manifest of supplementary materials package: MultimediaAppendix}} \\
        \toprule
        Path & Description  \\
        \midrule
        \verb|example_data/| & Example of audio records\\
        \quad \verb|radio_corpus/| & 5 West African Radio Corpus file samples\\
        \quad \verb|asr/| & 10 West African Virtual Assistant ASR Corpus file samples\\
        \quad \verb|lang_id/| & \\
        \quad \quad \verb|maninka/| & 2 Examples of audio records contaning Maninka language\\
        \quad \quad \verb|pular/| & Examples of audio records contaning Pular language\\
        \quad \quad \verb|susu/| & Examples of audio records contaning Susu language\\
    \bottomrule

    \end{tabular}
    \caption{Manifest ofMultimediaAppendix}
    \label{tab:manifest_multimedia_appendix}
\end{table}

\begin{table}
    \centering
    \begin{tabular}{ll}
        \multicolumn{2}{c}{\textbf{Manifest of supplementary materials package: TechnicalAppendix}} \\
        \toprule
        Path & Description  \\
        \midrule
        \verb|nicolingua_supplemental_materials.pdf| & this file\\
    \bottomrule

    \end{tabular}
    \caption{manifest of TechnicalAppendix}
    \label{tab:manifest_technical_appendix}
\end{table}

\FloatBarrier
\pagebreak
\section{Appendix B: Detailed Description of West African Virtual Assistant Speech Recognition Corpus}
\pagebreak

\begin{figure}
    \centering
    \includegraphics[width=\linewidth]{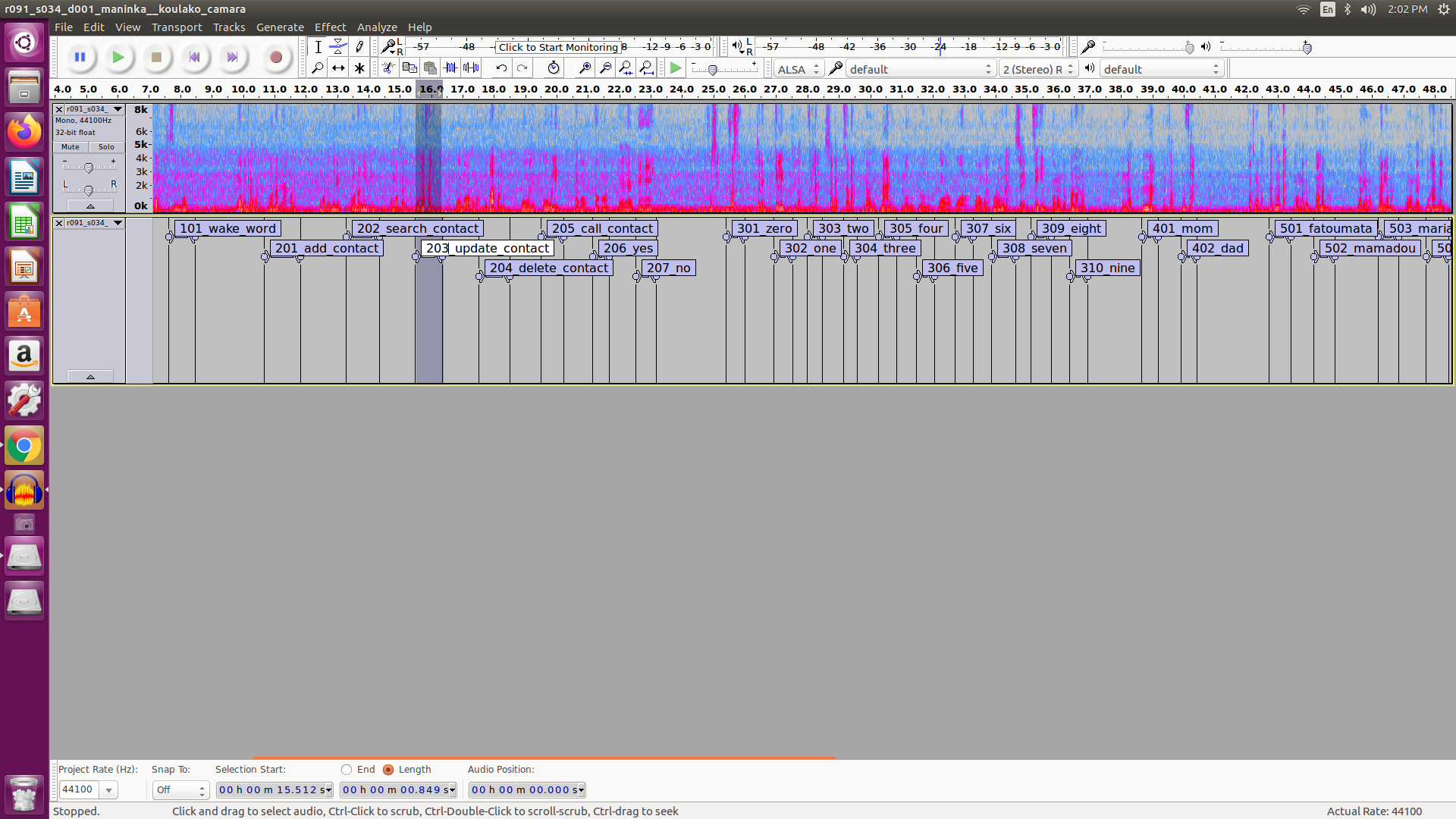}
    \caption{West African Virtual Assistant Speech Recognition Corpus: Annotation Process}
    \label{fig:asr_corpus_annotation}
\end{figure}

\begin{table}
    \centering
    \resizebox{0.25\linewidth}{!}{
    \begin{tabular}{rlr}
        \toprule
        Class ID & Utterance & Count \\ 
        \midrule
        0 & 101\_wake\_word\_\_francais & 66 \\ 
        1 & 101\_wake\_word\_\_maninka & 95 \\ 
        2 & 101\_wake\_word\_\_pular & 67 \\ 
        3 & 101\_wake\_word\_\_susu & 109 \\ 
        \midrule
        4 & 201\_add\_contact\_\_francais & 66 \\ 
        5 & 201\_add\_contact\_\_maninka & 95 \\ 
        6 & 201\_add\_contact\_\_pular & 67 \\ 
        7 & 201\_add\_contact\_\_susu & 111 \\ 
        \midrule
        8 & 202\_search\_contact\_\_francais & 66 \\ 
        9 & 202\_search\_contact\_\_maninka & 95 \\ 
        10 & 202\_search\_contact\_\_pular & 67 \\ 
        11 & 202\_search\_contact\_\_susu & 111 \\ 
        \midrule
        12 & 203\_update\_contact\_\_francais & 66 \\ 
        13 & 203\_update\_contact\_\_maninka & 95 \\ 
        14 & 203\_update\_contact\_\_pular & 67 \\ 
        15 & 203\_update\_contact\_\_susu & 111 \\ 
        \midrule
        16 & 204\_delete\_contact\_\_francais & 66 \\ 
        17 & 204\_delete\_contact\_\_maninka & 95 \\ 
        18 & 204\_delete\_contact\_\_pular & 67 \\ 
        19 & 204\_delete\_contact\_\_susu & 111 \\ 
        \midrule
        20 & 205\_call\_contact\_\_francais & 66 \\ 
        21 & 205\_call\_contact\_\_maninka & 95 \\ 
        22 & 205\_call\_contact\_\_pular & 64 \\ 
        23 & 205\_call\_contact\_\_susu & 111 \\ 
        \midrule
        24 & 206\_yes\_\_francais & 66 \\ 
        25 & 206\_yes\_\_maninka & 95 \\ 
        26 & 206\_yes\_\_pular & 67 \\ 
        27 & 206\_yes\_\_susu & 111 \\ 
        \midrule
        28 & 207\_no\_\_francais & 66 \\ 
        29 & 207\_no\_\_maninka & 95 \\ 
        30 & 207\_no\_\_pular & 67 \\ 
        31 & 207\_no\_\_susu & 111 \\ 
        \midrule
        32 & 301\_zero\_\_francais & 66 \\ 
        33 & 301\_zero\_\_maninka & 95 \\ 
        34 & 301\_zero\_\_pular & 67 \\ 
        35 & 301\_zero\_\_susu & 111 \\ 
        36 & 302\_one\_\_francais & 66 \\ 
        37 & 302\_one\_\_maninka & 95 \\ 
        38 & 302\_one\_\_pular & 67 \\ 
        39 & 302\_one\_\_susu & 111 \\ 
        40 & 303\_two\_\_francais & 66 \\ 
        41 & 303\_two\_\_maninka & 95 \\ 
        42 & 303\_two\_\_pular & 67 \\ 
        43 & 303\_two\_\_susu & 111 \\ 
        44 & 304\_three\_\_francais & 66 \\ 
        45 & 304\_three\_\_maninka & 95 \\ 
        46 & 304\_three\_\_pular & 67 \\ 
        47 & 304\_three\_\_susu & 111 \\ 
        48 & 305\_four\_\_francais & 66 \\ 
        49 & 305\_four\_\_maninka & 95 \\ 
        50 & 305\_four\_\_pular & 67 \\ 
        51 & 305\_four\_\_susu & 111 \\ 
        52 & 306\_five\_\_francais & 66 \\ 
        53 & 306\_five\_\_maninka & 95 \\ 
        54 & 306\_five\_\_pular & 67 \\ 
        55 & 306\_five\_\_susu & 111 \\ 
        56 & 307\_six\_\_francais & 66 \\ 
        57 & 307\_six\_\_maninka & 95 \\ 
        58 & 307\_six\_\_pular & 67 \\ 
        59 & 307\_six\_\_susu & 111 \\ 
        60 & 308\_seven\_\_francais & 66 \\ 
        61 & 308\_seven\_\_maninka & 95 \\ 
        62 & 308\_seven\_\_pular & 67 \\ 
        63 & 308\_seven\_\_susu & 111 \\ 
        64 & 309\_eight\_\_francais & 66 \\ 
        65 & 309\_eight\_\_maninka & 93 \\ 
        66 & 309\_eight\_\_pular & 67 \\ 
        67 & 309\_eight\_\_susu & 111 \\ 
        68 & 310\_nine\_\_francais & 66 \\ 
        69 & 310\_nine\_\_maninka & 93 \\ 
        70 & 310\_nine\_\_pular & 67 \\ 
        71 & 310\_nine\_\_susu & 111 \\ 
        \midrule
        72 & 401\_mom\_\_francais & 36 \\ 
        73 & 401\_mom\_\_maninka & 53 \\ 
        74 & 401\_mom\_\_pular & 43 \\ 
        75 & 401\_mom\_\_susu & 51 \\ 
        76 & 402\_dad\_\_francais & 36 \\ 
        77 & 402\_dad\_\_maninka & 53 \\ 
        78 & 402\_dad\_\_pular & 43 \\ 
        79 & 402\_dad\_\_susu & 51 \\ 
        \midrule
        80 & 501\_fatoumata\_\_\_language\_independent & 146 \\ 
        81 & 502\_mamadou\_\_\_language\_independent & 145 \\ 
        82 & 503\_mariama\_\_\_language\_independent & 145 \\ 
        83 & 504\_mohamed\_\_\_language\_independent & 145 \\ 
        84 & 505\_kadiatou\_\_\_language\_independent & 145 \\ 
        85 & 506\_ibrahima\_\_\_language\_independent & 145 \\ 
        86 & 507\_aissatou\_\_\_language\_independent & 145 \\ 
        87 & 508\_aminata\_\_\_language\_independent & 145 \\ 
        88 & 509\_alpha\_\_\_language\_independent & 145 \\ 
        89 & 510\_thierno\_\_\_language\_independent & 142 \\ 
        90 & 511\_abdoulaye\_\_\_language\_independent & 145 \\ 
        91 & 512\_aboubacar\_\_\_language\_independent & 145 \\ 
        92 & 513\_amadou\_\_\_language\_independent & 145 \\ 
        93 & 514\_fanta\_\_\_language\_independent & 145 \\ 
        94 & 515\_mariame\_\_\_language\_independent & 145 \\ 
        95 & 516\_oumou\_\_\_language\_independent & 145 \\ 
        96 & 517\_ousmane\_\_\_language\_independent & 145 \\ 
        97 & 518\_adama\_\_\_language\_independent & 145 \\ 
        98 & 519\_marie\_\_\_language\_independent & 145 \\ 
        99 & 520\_moussa\_\_\_language\_independent & 145 \\ 
        100 & 521\_aissata\_\_\_language\_independent & 145 \\ 
        101 & 522\_hawa\_\_\_language\_independent & 145 \\ 
        102 & 523\_sekou\_\_\_language\_independent & 145 \\ 
        103 & 524\_hadja\_\_\_language\_independent & 145 \\ 
        104 & 525\_djenabou\_\_\_language\_independent & 146 \\ 
        \bottomrule
    \end{tabular}
    }
    \caption{West African Virtual Assistant Speech Recognition Corpus: Records by class}
\end{table}

\begin{table}
    \centering
    \begin{tabular}{lrrrr}
    \toprule
        \multicolumn{1}{c}{Utterance} & 
        \multicolumn{1}{c}{French} & 
        \multicolumn{1}{c}{Maninka} & 
        \multicolumn{1}{c}{Pular} & 
        \multicolumn{1}{c}{Susu} \\
        \midrule
        101 - Wake word & 66 & 95 & 67 & 109 \\
        201 - Add  & 66 & 95 & 67 & 111 \\
        202 - Search & 66 & 95 & 67 & 111 \\
        203 - Update & 66 & 95 & 67 & 111 \\
        204 - Delete & 66 & 95 & 67 & 111 \\
        205 - Call & 66 & 95 & 64 & 111 \\
        206 - Yes & 66 & 95 & 67 & 111 \\
        207 - No & 66 & 95 & 67 & 111 \\
        301 - 310 Digits & 660 & 946 & 670 & 1,110 \\
        401 - Mom & 36 & 53 & 43 & 51 \\
        402 - Dad & 36 & 53 & 43 & 51 \\ \midrule
        \textbf{Total/Language} & 1,260 & 1,812 & 1,289 & 2,098 \\ \midrule
        501 - 525 Names & \multicolumn{4}{c}{3,624} \\ \midrule
        \textbf{Total} & \multicolumn{4}{c}{10,083} \\
    \bottomrule
    \end{tabular}
    
    \caption{West African Virtual Assistant Speech recognition corpus. Records by utterance category and language}
    \label{table:asr_corpus_stats}
\end{table}

\begin{table}
    \centering
    \begin{tabular}{ccr}
    \toprule
        Device ID & Device Model & Record Count \\ \midrule
    	d001 & Macbook Pro & 2759 \\
    	d002 & Google Pixel 2 & 2759 \\
    	d003 & Huawei P30 Pro & 2759 \\
    	d004 & Samsung sm n900v & 762 \\
    	d005 & Redmi 8a & 422 \\
    	d006 & Tekno lb6 & 320 \\
    	d007 & Oneplus 6t & 65 \\
    	d008 & Samsung s9 & 65 \\
    	d009 & Samsung a30 & 172 \\ \midrule
        \textbf{Total} & \multicolumn{2}{c}{10,083} \\
        \bottomrule
    \end{tabular}
    
    \caption{West African Virtual Assistant Speech recognition corpus. Records by recording device}
    \label{table:asr_corpus_stats_by_device}
\end{table}

\begin{table}
    \centering
    \begin{tabular}{cc}
    \toprule
        Gender & Count  \\ \midrule
        Female & 3440  \\
        Male & 6643 \\ \midrule
        \textbf{Total} & 10083 \\
        \bottomrule

    \end{tabular}
    \caption{West African Virtual Assistant Speech recognition corpus. Records by speaker Gender}
    \label{tab:asr_corpus_stats_by_gender}
\end{table}

\begin{table}
    \centering
    \begin{tabular}{cc}
    \toprule
        In Speaker's mother tongue & Count \\ \midrule
        False & 7698 \\
        True & 2385 \\ \midrule
        \textbf{Total}   & 10083 \\
        \bottomrule
    \end{tabular}
    \caption{West African Virtual Assistant Speech recognition corpus. Records by whether utterances were recorded in the mother tongue of the speaker. First names are in the 'False' category}
    \label{tab:asr_corpus_stats_by_spoken_in_mothertongue}
\end{table}

\begin{table}
    \centering
    \resizebox{\columnwidth}{!}{
    \begin{tabular}{ll}
    \toprule
        \multicolumn{1}{c}{Field} & 
        \multicolumn{1}{c}{Description} \\ \midrule
        file & Audio file name \\
        recording\_session\_id & Recording session identifier \\
        speaker\_id & Speaker Identifier \\
        device\_id & Recording Device Identifier \\
        language & Language in which the recording was done \\
        utterance\_id & uterance Category \\
        label & ASR class id. Unique across (Utterance\_id, language) \\
        speaker\_age & Age of the speaker \\
        speaker\_gender & Gender of the speaker \\
        speaker\_mothertongue & Mothergongue of the Speaker \\
        \bottomrule
    \end{tabular}
    }
    \caption{West African Virtual Assistant Speech recognition corpus. Fields of the metadata file}
    \label{tab:asr_metadata_description}
\end{table}

\begin{figure}
    \centering
    \includegraphics[width=0.9\columnwidth]{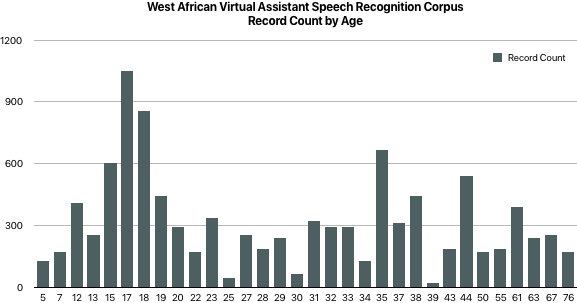}
    \caption{VA ASR Corpus: Record Count by Age}
    \label{fig:my_label}
\end{figure}

\FloatBarrier
\pagebreak
\section{Appendix C: Detailed description of West African Radio Corpus}
\pagebreak

\begin{figure}
    \centering
    \includegraphics[width=\linewidth]{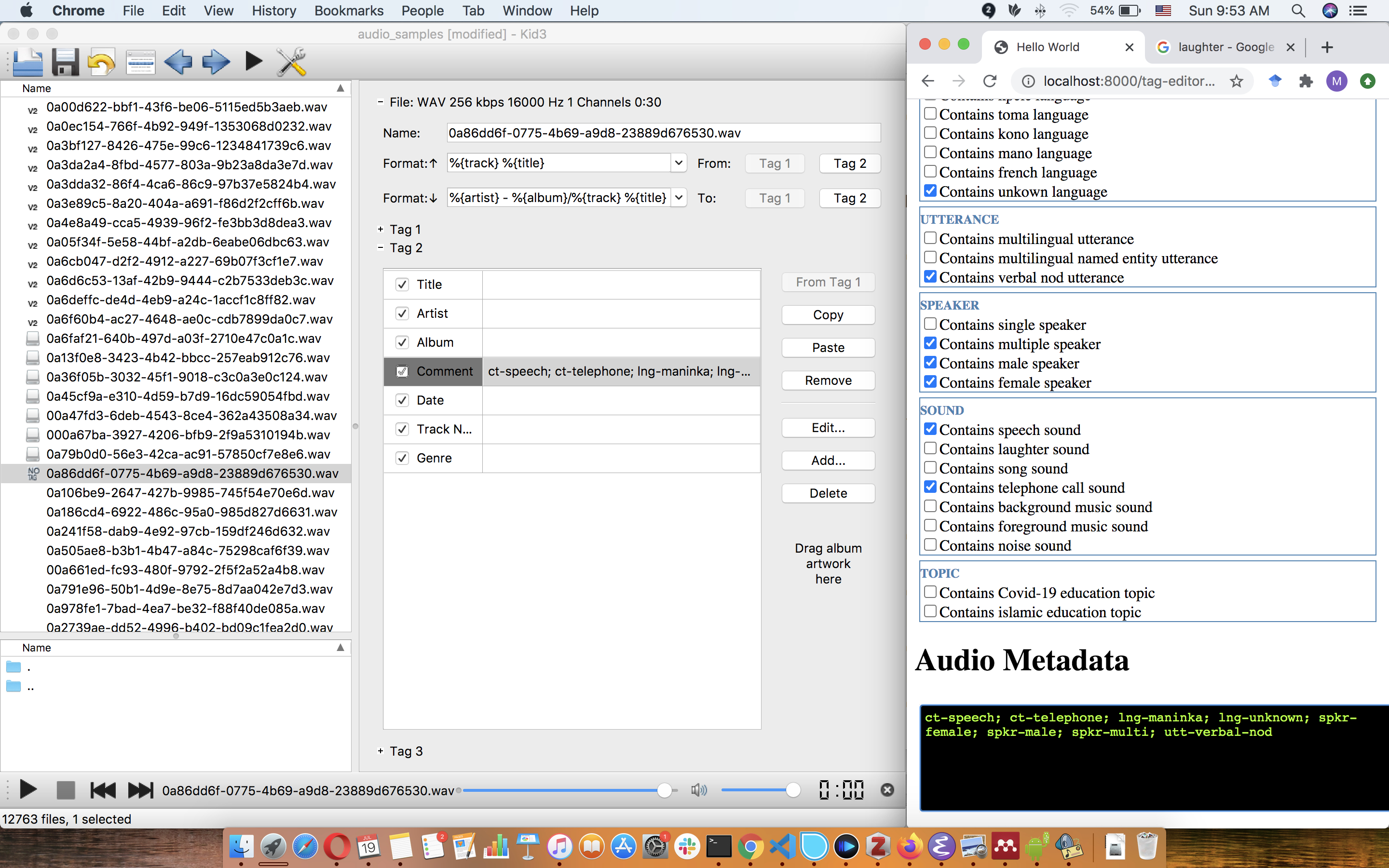}
    \caption{Radio Corpus Validation Set: Annotation Process. Kid3 was used to store tags in the ID3v2 metadata. Scripyt was subsequently used to retrieve the tags and build a metadata file.}
    \label{fig:radio_corpus_annotation_process}
\end{figure}

\begin{figure}
\centering
\includegraphics[width=0.9\columnwidth]{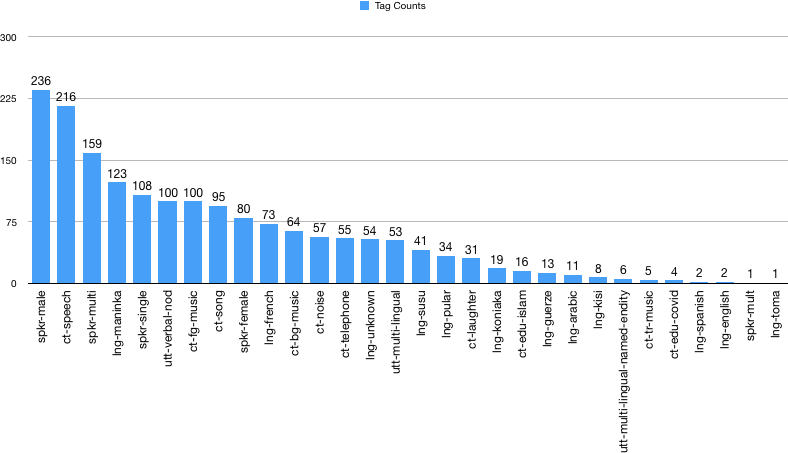} 
\caption{Frequency of the tags in the test set of the West African Radio Corpus Validation Set.}.
\label{fig:radio_test_tag_freq}
\end{figure}

\FloatBarrier
\pagebreak
\section{Appendix D: Detailed results for Language Identification Experiments}
\pagebreak

\begin{figure}
\centering
\includegraphics[width=0.9\columnwidth]{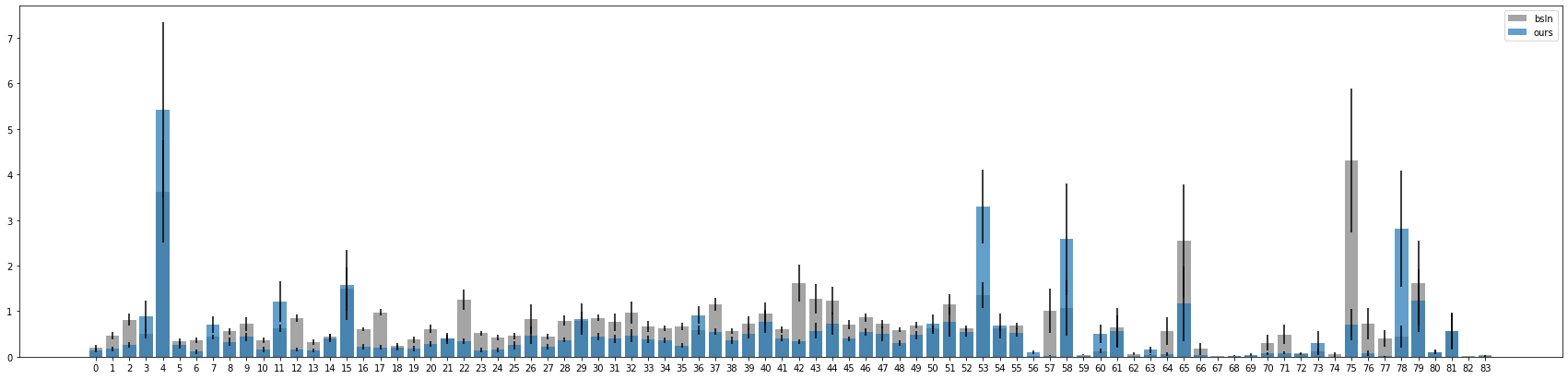} 
\caption{Mean and Standard Error of KL-Divergences between the true class distribution and the language identification network predictions  for  each example of the language identification dataset across 10 models resulting from 10 cross validation splits using the baseline wav2vec(grey), and the west african wav2vec (blue)}
\label{fig:lang_id_kl_divergences}
\end{figure}

\begin{figure}
\centering
\includegraphics[width=0.9\columnwidth]{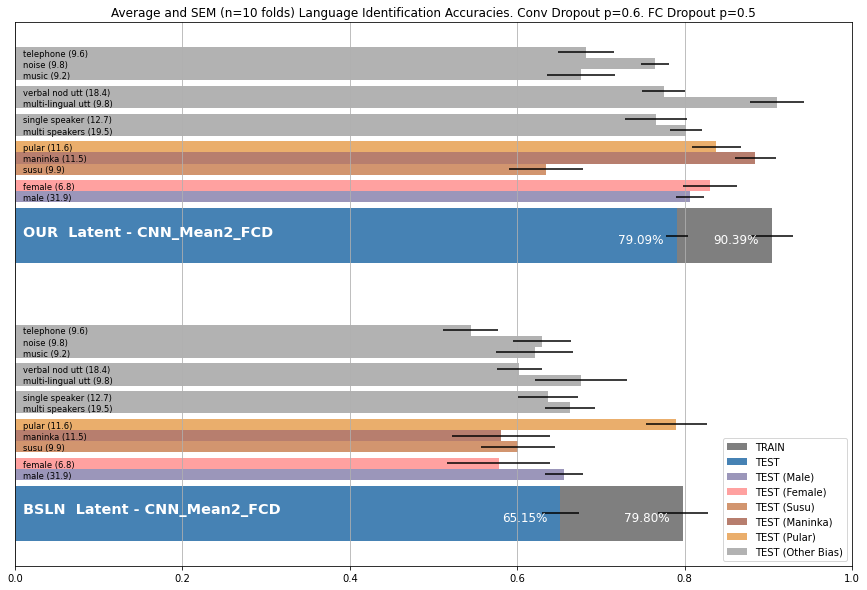}
\caption{Language Identification Results: Latent wav2vec features}.
\label{fig:langid_results_summary_latent}
\end{figure}

\begin{figure}
\centering
\includegraphics[width=0.9\columnwidth]{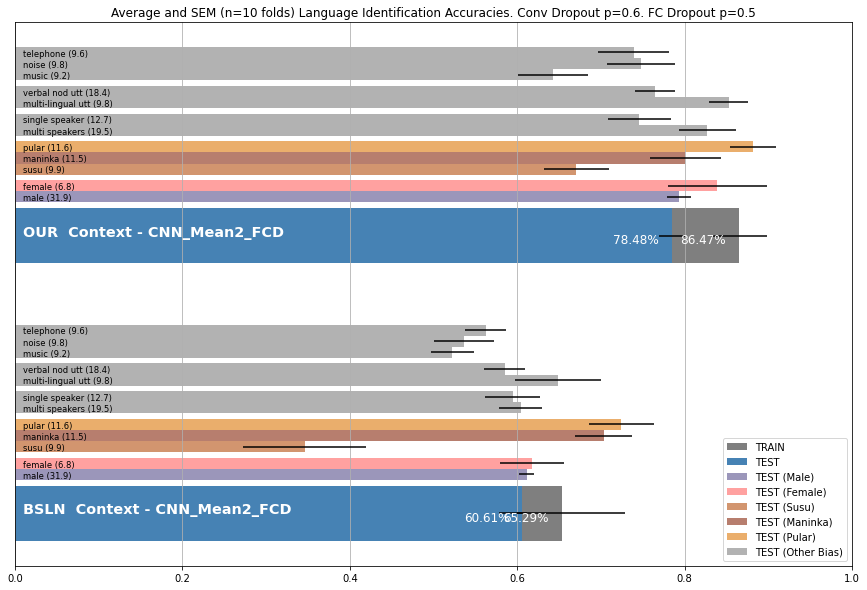}
\caption{Language Identification Results: Context wav2vec features}.
\label{fig:langid_results_summary_context}
\end{figure}

\begin{figure}
\centering
0f4c3c2e-5e44-4c7a-a2b7-95799919ee07.wav \\
\includegraphics[width=0.9\linewidth]{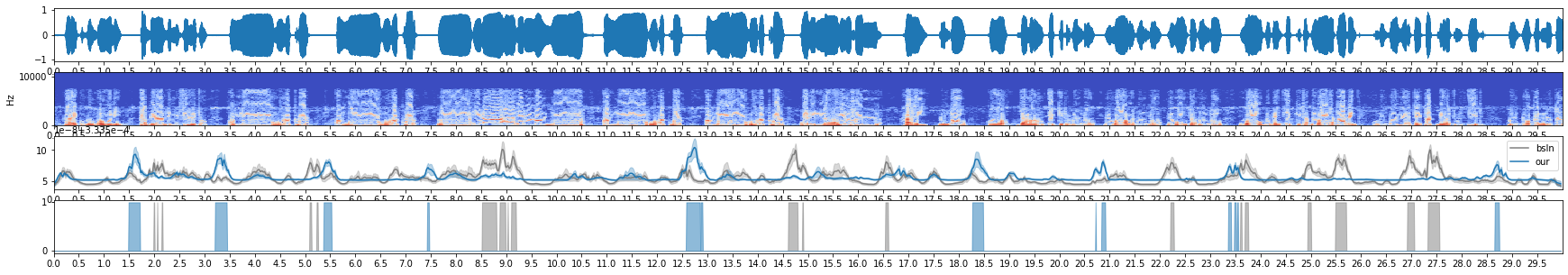} \\
0c73a83d-c0fc-4ac4-994c-a2b14ede31cb.wav \\
\includegraphics[width=0.9\linewidth]{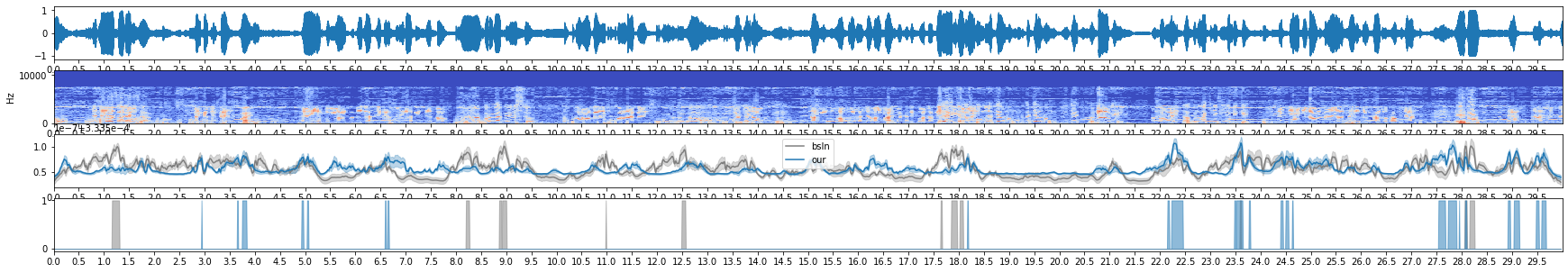} \\
0c424e24-ac12-45d6-a830-4c828b0db2ec.wav \\
\includegraphics[width=0.9\linewidth]{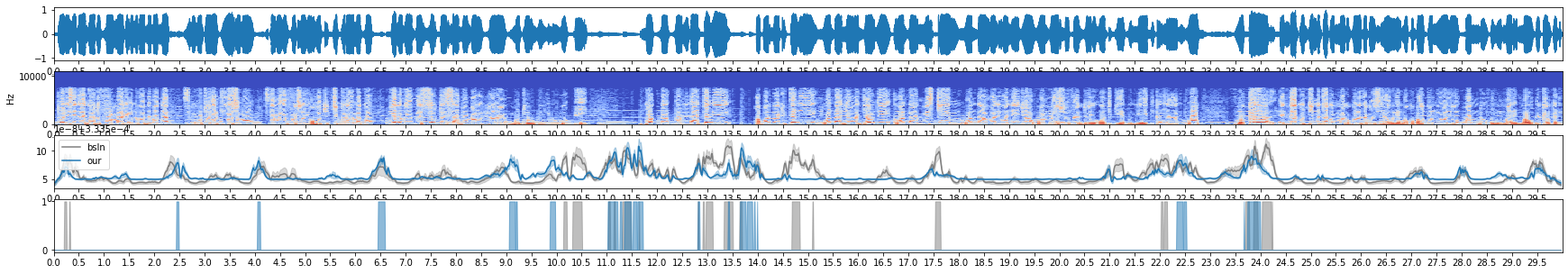} \\
0cec2f3a-e739-47bd-9fc1-c9d8431c38db.wav \\
\includegraphics[width=0.9\linewidth]{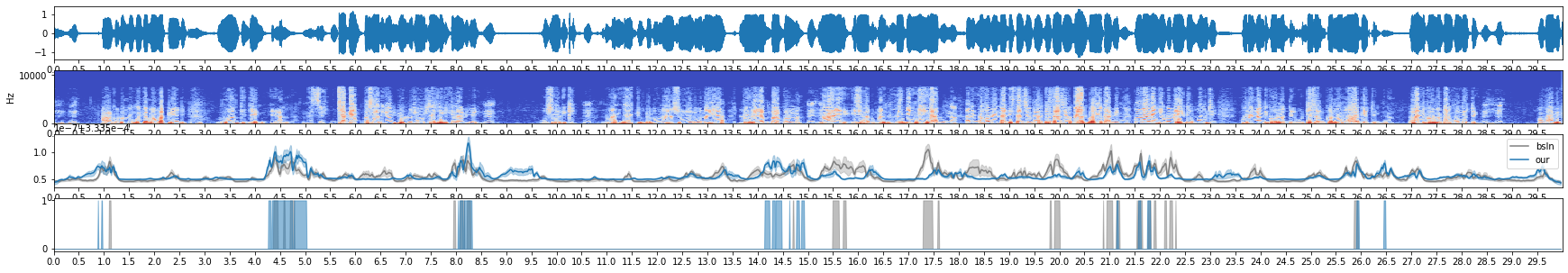} \\
0f510424-6f24-4b02-8554-3c7f0a40fadf.wav \\
\includegraphics[width=0.9\linewidth]{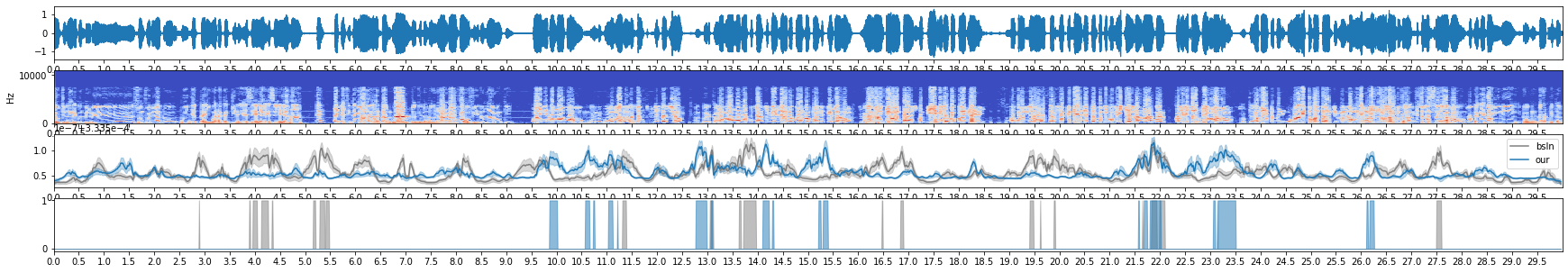}

\caption{Example of audio files with attention signals and segmented acoustic units. The corresponding audio files and audacity markers (in a labels file) are provided in the CodeAndDataAppendix. See Table \ref{table:manifest_code_and_data_appendix} for details.
}
\label{fig:accoustic_units_segmentation}
\end{figure}

\begin{figure}
\centering
\includegraphics[width=0.45\linewidth]{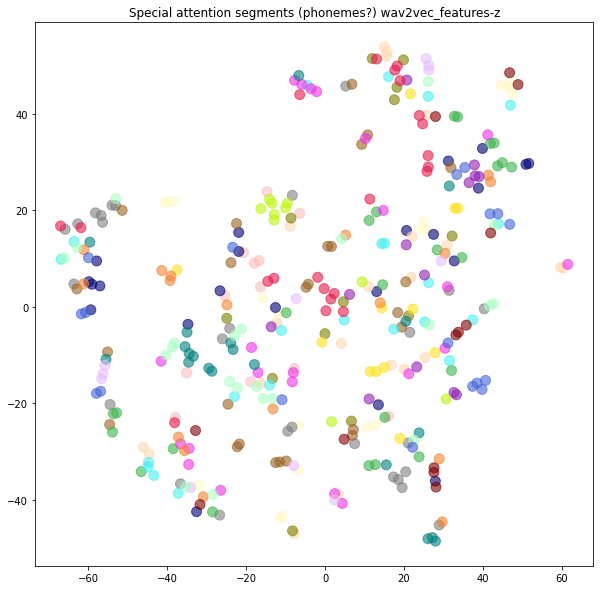}
\includegraphics[width=0.45\linewidth]{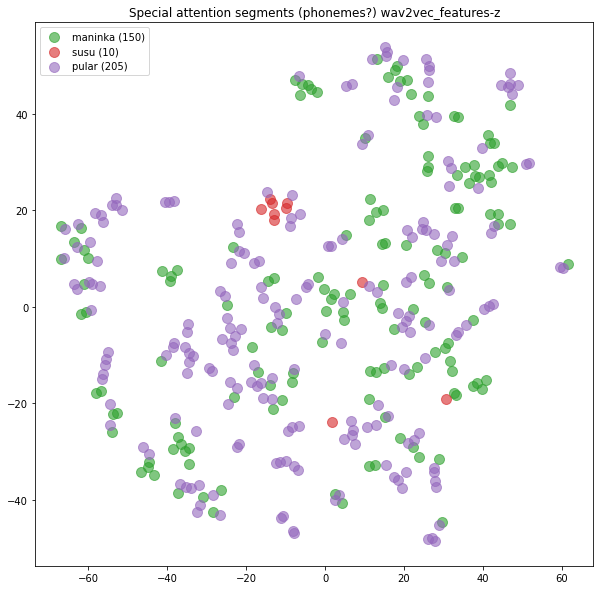} \\
\includegraphics[width=0.45\linewidth]{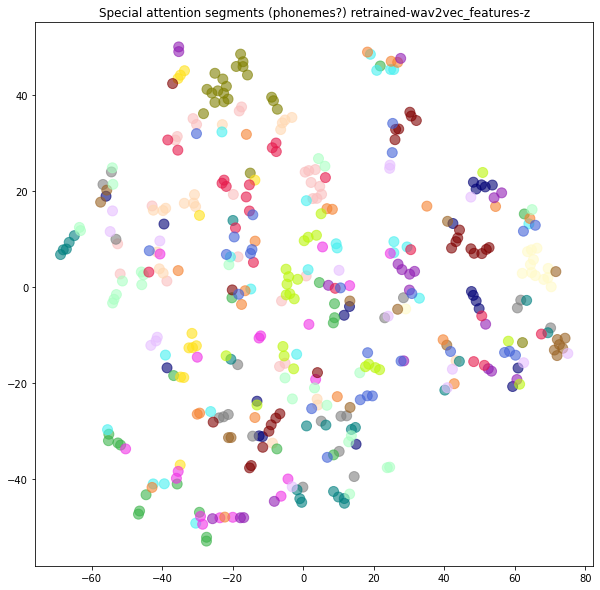}
\includegraphics[width=0.45\linewidth]{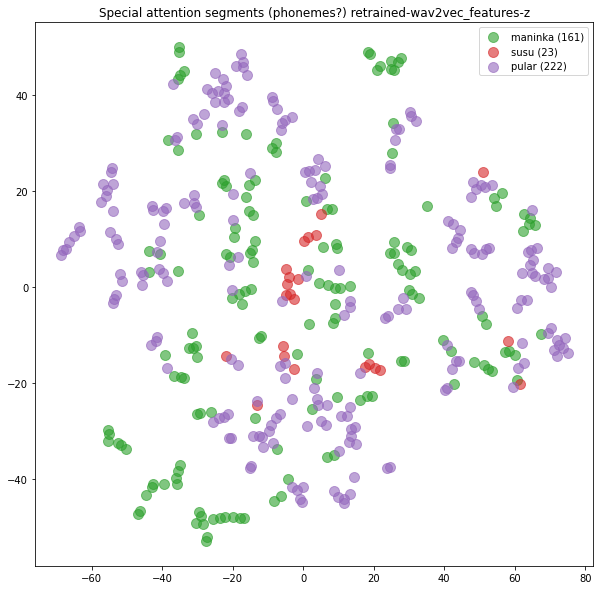}

\caption{T-SNE projection of segmented acoustic units}.
\label{fig:segmented_acoustic_units}
\end{figure}

\begin{figure}
\centering
\includegraphics[width=0.3\linewidth]{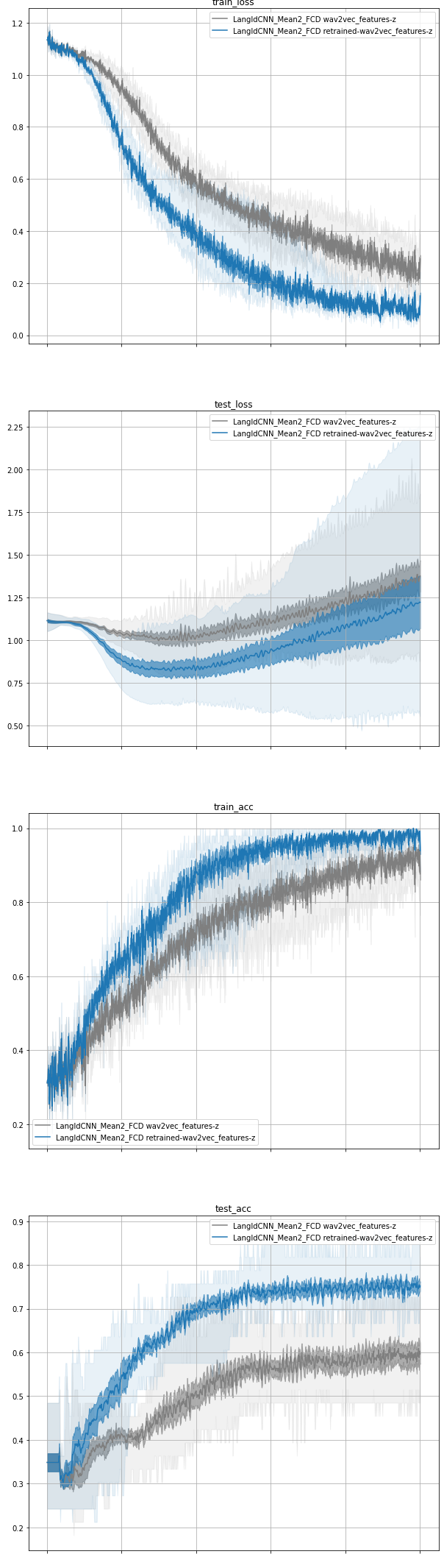}
\caption{Training details for Languge ID}.
\label{fig:asr_taining_details_ctx}
\end{figure}

\FloatBarrier
\pagebreak
\section{Appendix E: Detailed results for Speech Recognition Experiments}
\pagebreak

\begin{figure}
\centering
\includegraphics[width=0.9\columnwidth]{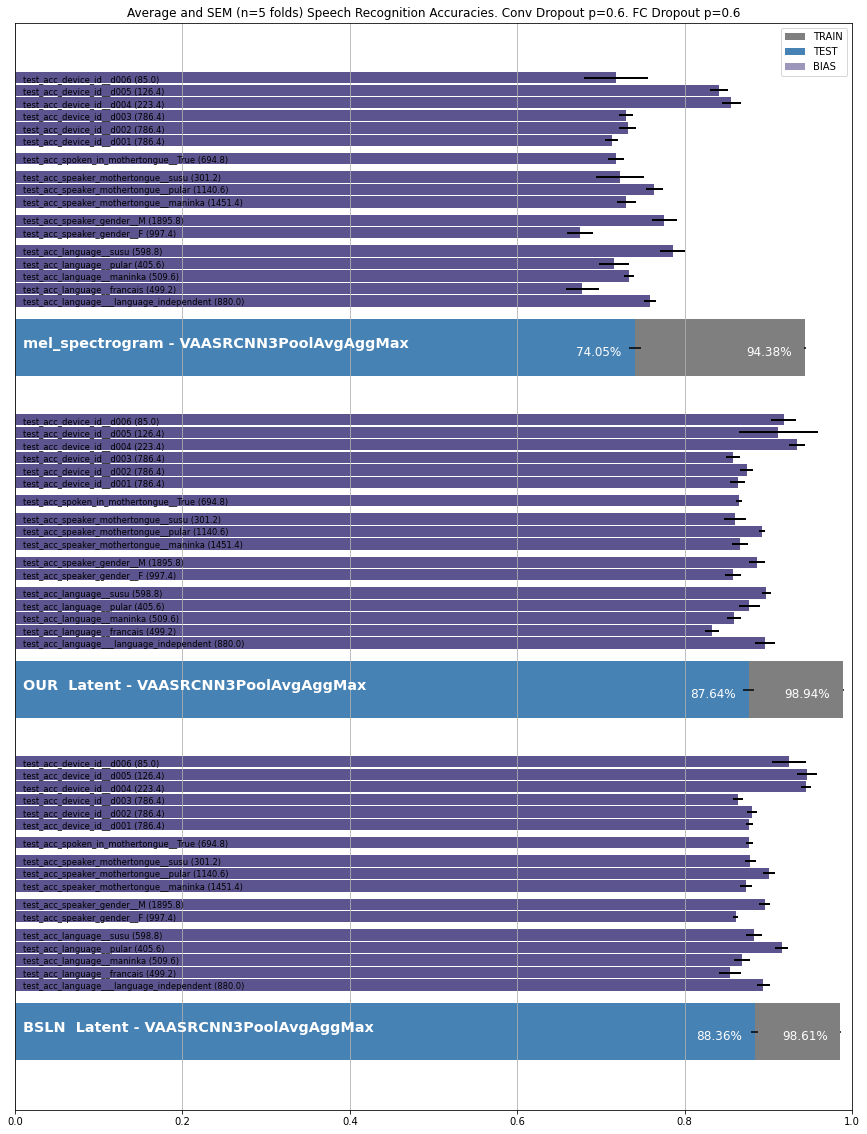}
\caption{ASR Results: Latent wav2vec features}.
\label{fig:asr_results_summary_latent}
\end{figure}

\begin{figure}
\centering
\includegraphics[width=0.9\columnwidth]{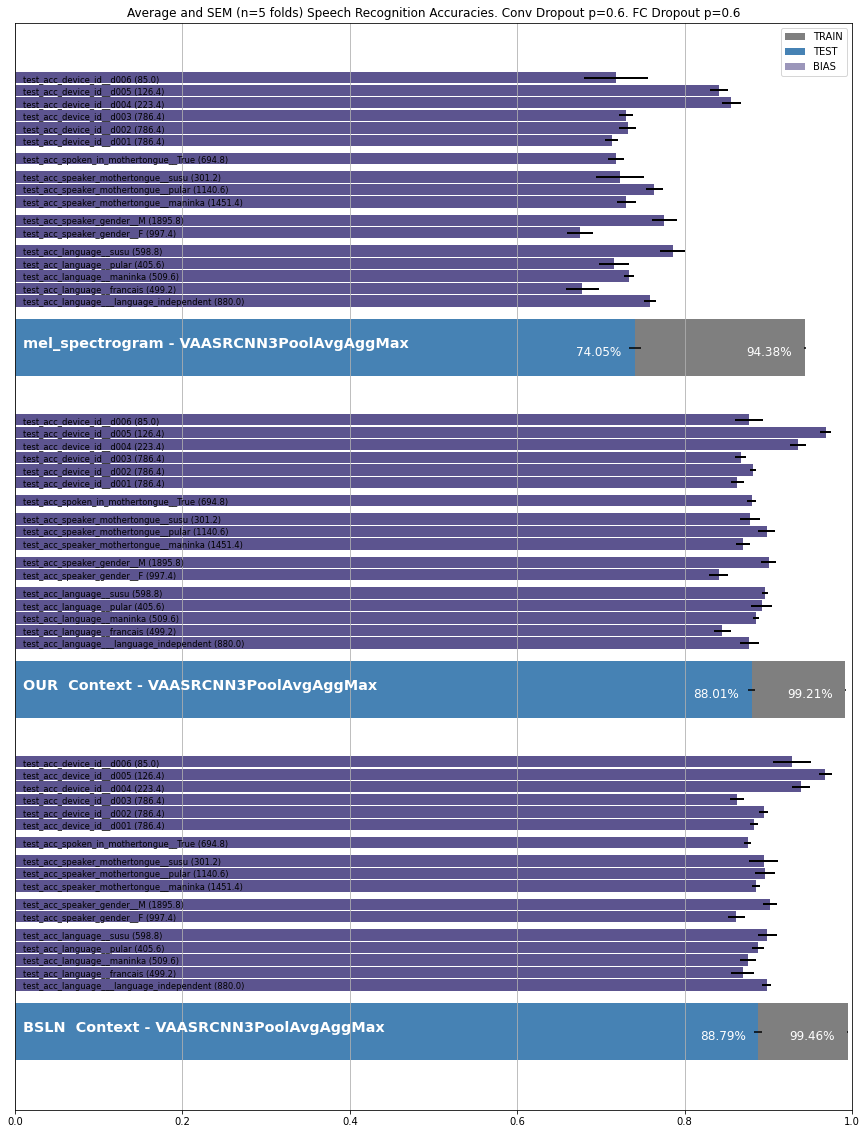}
\caption{ASR Results: Context Wav2vec features}.
\label{fig:asr_results_summary_context}
\end{figure}

\begin{figure}
\centering
\includegraphics[width=0.3\linewidth]{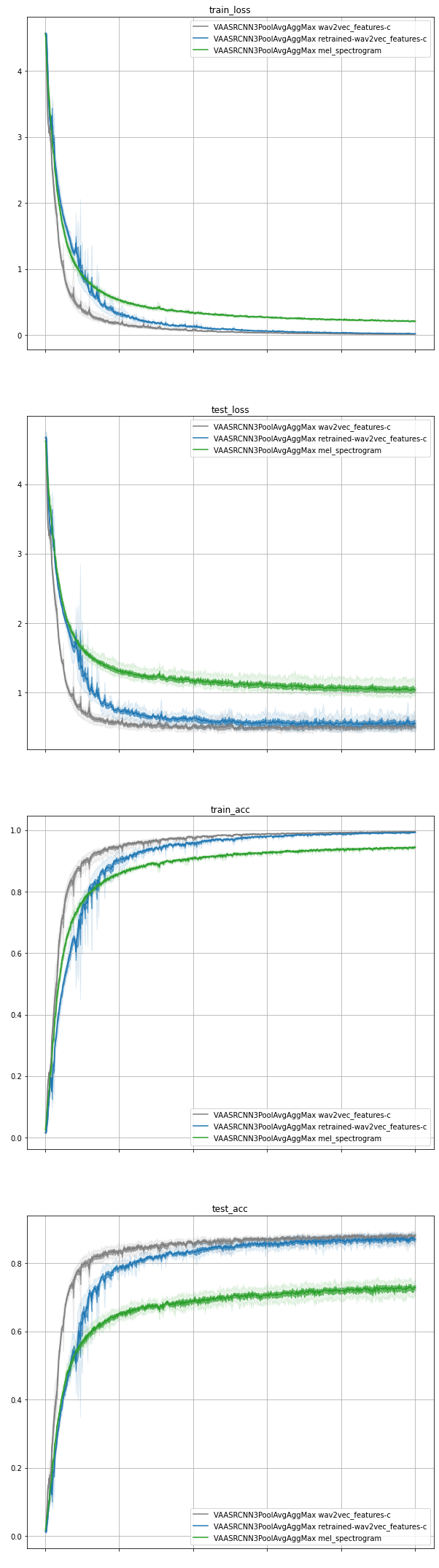}
\caption{ASR Learning Curves: Context Wav2vec Features}.
\label{fig:asr_taining_details_context}
\end{figure}

\begin{figure}
\centering
\includegraphics[width=0.3\columnwidth]{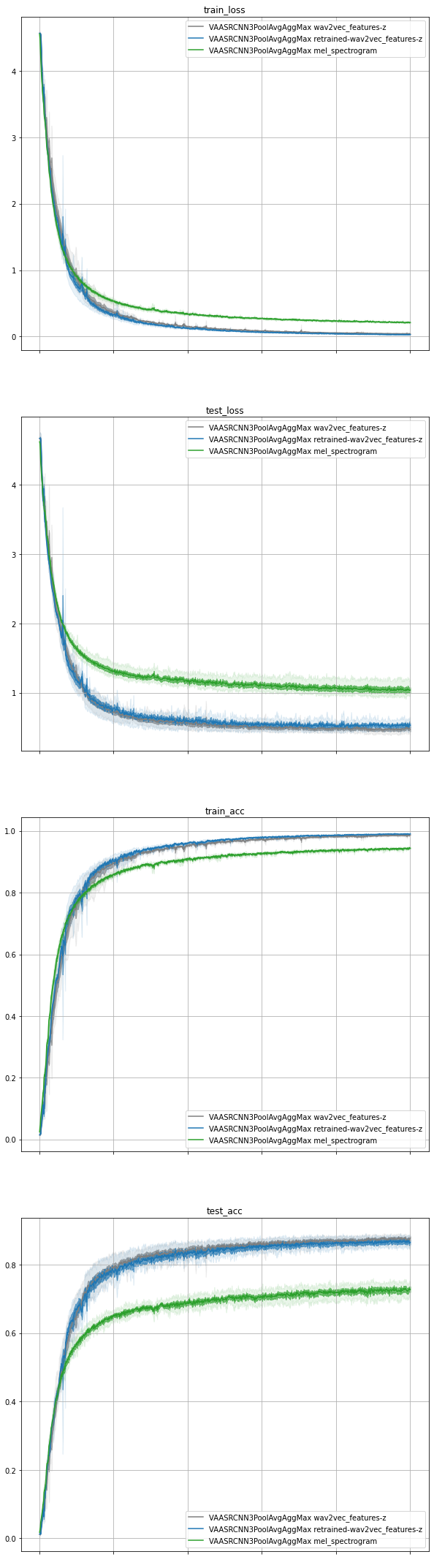}
\caption{ASR Learning Curves: Context Wav2vec Features}.
\label{fig:asr_taining_details_latent}
\end{figure}
